\documentclass{article}

 \usepackage[main, final]{neurips_2026}

\usepackage[utf8]{inputenc} 
\usepackage[T1]{fontenc}    
\usepackage{hyperref}       
\usepackage{url}            
\usepackage{booktabs}       
\usepackage{amsfonts}       
\usepackage{nicefrac}       
\usepackage{microtype}      
\usepackage{xcolor}         

\usepackage{amsmath}
\usepackage{amssymb}
\usepackage{mathtools}
\usepackage{amsthm}

\usepackage{multirow}
\usepackage{pifont}
\usepackage{makecell}
\usepackage{xcolor}
\usepackage{tcolorbox}
\tcbuselibrary{breakable,skins}
\usepackage{wrapfig}
\usepackage{enumitem}

\usepackage[capitalize,noabbrev]{cleveref}

\theoremstyle{plain}
\newtheorem{theorem}{Theorem}[section]

\theoremstyle{definition}
\newtheorem{definition}[theorem]{Definition}

\theoremstyle{remark}
\newtheorem{remark}[theorem]{Remark}

\title{HierFlow: Hierarchical Coupled Dual-Space Search for Automatic Agentic Workflow Generation}

\newcommand{\sysname}{HierFlow}

\author{
  Dong Li$^{1}$ \quad Yanchi Liu$^{2*}$ \quad Xujiang Zhao$^{3}$ \quad Wei Cheng$^{2}$ \quad 
  \textbf{Zhengzhang Chen}$^{2}$ \\ \textbf{Xintao Wu}$^{4}$ \quad \textbf{Zhong Chen}$^{5}$ \quad \textbf{Haifeng Chen}$^{2*}$ \quad \textbf{Chen Zhao}$^{1}$\thanks{Corresponding authors.} 
  \\
  $^{1}$Baylor University \quad
  $^{2}$NEC Labs America \quad
  $^{3}$Meta\quad
  $^{4}$University of Arkansas \\
  $^{5}$Southern Illinois University
  \\
  \texttt{\{dong\_li1, chen\_zhao\}@baylor.edu} \quad
  \texttt{\{yanchi, Haifeng\}@nec-labs.com} \\
}

\begin{document}

\maketitle

\begin{abstract}
Agentic AI systems enable LLMs to solve non-trivial tasks through structured workflows, but automatically generating such workflows remains challenging due to the discrete and combinatorial search space. Existing methods often rely on offline search or training, limiting query-level adaptability and incurring substantial data and engineering overhead. We formulate workflow generation as a coupled topology--execution search problem, where the upper-level topology induces subtask-specific code-search domains and lower-level execution feedback can revise the topology itself. Based on this formulation, we propose HierFlow, a training-free hierarchical test-time search framework for automatic agentic workflow generation. HierFlow couples feedback-driven topology refinement with an MCTS-inspired lightweight tree search for execution-level sub-workflow optimization, and uses an adaptive gating mechanism to selectively trigger execution-level search based on estimated necessity. We further provide a coupling-aware analysis characterizing when hierarchical decomposition and proxy-based gating are beneficial and when their advantages may degrade under stronger cross-subtask coupling. Experiments across QA, mathematical reasoning, and code generation benchmarks show that HierFlow achieves strong performance and favorable efficiency--quality trade-offs, outperforming competitive baselines without additional training. 
\end{abstract}

\section{Introduction}
\label{sec:introduction}
\begin{wrapfigure}{R}{0.65\textwidth}
    \centering
    \includegraphics[width=\linewidth]{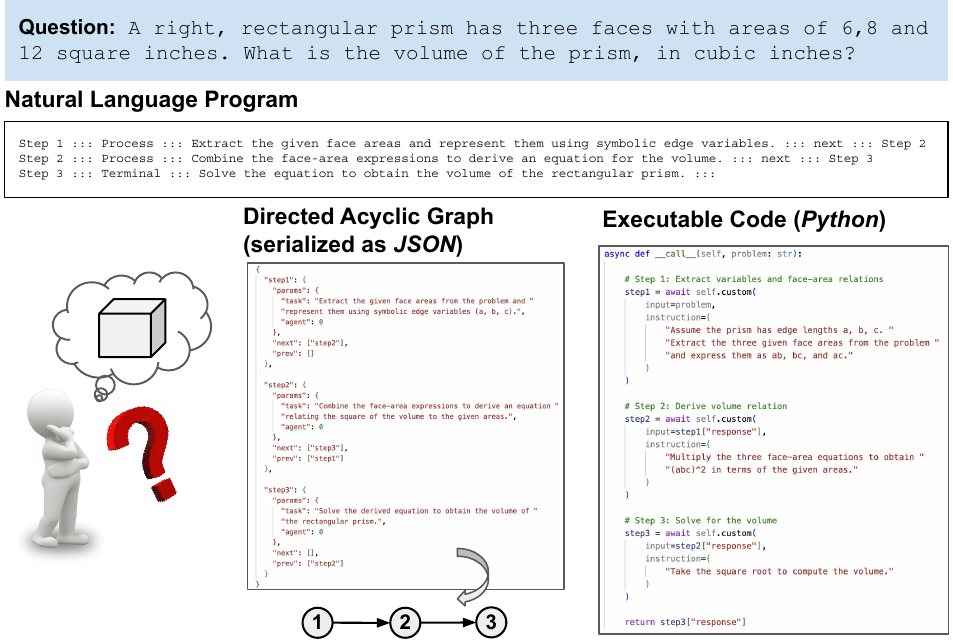}
    \caption{
    Three equivalent workflow representations (proxies) for a simple mathematical question.
    }
    \label{fig:intro}
\end{wrapfigure}
Recent advances in large language models (LLMs) have led to the emergence of agentic AI systems~\cite{hughes2025ai} that solve non-trivial tasks through multi-step reasoning and action~\cite{acharya2025agentic,li2024survey}. In such systems, workflows provide a structured way to organize task decomposition, intermediate reasoning, and interactions with tools or other agents~\cite{singh2024enhancing}.
Early agentic reasoning methods predominantly rely on handcrafted workflows~\cite{CoT,yao2023tree,besta2024graph}, where the reasoning structure and execution procedure are manually specified and fixed at inference time, such as Self-Refine~\cite{SR}, and LLM-Debate~\cite{llmdebate}.
Despite their effectiveness, these methods share a common limitation: the workflow is manually designed and remains static, limiting adaptability as task complexity increases~\cite{xu2025robustflow}.

A growing body of work in this area has focused on automatic agentic workflow generation~\cite{aflow}, framing the problem as a search-based optimization task that aims to identify an effective workflow for a given task or query~\cite{survey}. Due to the abstract, discrete, and combinatorially large nature of the workflow space, direct search is intractable~\cite{yu2025survey,qiao2024benchmarking}; consequently, existing methods typically operate in a proxy space (\textit{e.g.,} code and topology) with a mapping to workflow semantics, where each proxy configuration corresponds to a possible workflow. For example, AFlow~\cite{aflow} adapts Monte Carlo Tree Search (MCTS)~\cite{MCTS} to explore task-specific workflows in a programmatic code space. Figure~\ref{fig:intro} shows that the same workflow can be equivalently represented using different proxy forms.

Although these methods surpass the performance of manually designed systems, they mainly generate task-level workflows through offline search~\cite{adas,aflow} or query-level workflows via training a designer~\cite{maas,dyflow}, which not only incur substantial data and engineering costs but also remain tied to fixed training distributions, thereby limiting their generalization ability.
To address these limitations, test-time scaling has emerged as a promising training-free and query-adaptive paradigm~\cite{li2025solverllm}.
However, existing methods typically restrict to a single proxy space, leading to efficiency challenges.
In practice, proxy  differ substantially in expressiveness, search efficiency, and structural constraints~\cite{survey}, which suggests that leveraging multiple proxy spaces during search can effectively exploit their complementary strengths.


To this end, we propose \sysname{}, a test-time hierarchical search framework over two coupled proxy spaces: an upper-level \emph{topology space} for task decomposition and a lower-level \emph{code execution space} for subtask implementation. \sysname{} performs feedback-driven topology refinement and MCTS-inspired execution-level search, where the topology induces subtask-specific code-search domains and lower-level execution feedback can revise the topology itself. This makes the search object a coupled structure--execution state rather than a fixed pipeline over a single workflow representation. To balance quality and cost, we introduce an adaptive gating mechanism that invokes lower-level search only when a search-necessity proxy indicates potential benefit. We further provide a coupling-aware analysis characterizing when hierarchical decomposition and proxy-based gating are beneficial, and when their advantages may degrade under stronger cross-subtask coupling. Our main contributions are summarized as follows:
\begin{itemize}[itemsep=0pt, topsep=0pt]
    \item We formulate query-level workflow generation as coupled topology--execution search, where the topology is a search variable that induces subtask-level code-search domains and execution feedback can revise the topology at test time.

\item We propose an adaptive gating mechanism for budget-aware cross-space coordination, selectively triggering subtask-level code search when the current topology and subtask uncertainty suggest additional execution-level refinement is useful.

    \item We provide a coupling-aware analysis of hierarchical workflow search, showing how observable dependency density, execution-triggered repair, and gating-proxy informativeness affect the expected efficiency and boundary behavior of HierFlow.
    
    \item We conduct comprehensive empirical evaluations across diverse benchmarks to demonstrate the effectiveness, deployment robustness without retraining, and efficiency--quality trade-offs of \sysname{}, and to validate its overall design.

\end{itemize}

\section{Related Work}
\label{sec:related}

\noindent \textbf{Workflow Representation via Proxy Spaces.} Since direct search over the workflow space is highly challenging, existing approaches typically reformulate workflow generation by introducing a proxy space, in which searching over proxy representations serves as a surrogate for searching over workflows themselves~\cite{survey}. A key property of a proxy space is that each proxy instance induces a valid workflow, enabling indirect optimization over the workflow space. Depending on the specific form of the proxy, existing methods commonly operate in natural language (NL) program~\cite{autoflow}, code~\cite{aflow}, topology~\cite{flow}, or handcrafted spaces~\cite{dyflow}, which offer different trade-offs between expressiveness and tractability. 
NL proxies are highly expressive but excessively redundant and thus difficult to search; code proxies provide more structured representations but remain complex due to execution-level combinatorics; topology proxies abstract away execution details, yielding a smaller but less expressive space suitable for coarse-grained pruning; and handcrafted proxies further restrict the space to improve tractability at the cost of flexibility.

\begin{wraptable}{r}{0.55\textwidth}
\caption{Comparison between automatic agentic workflow generation methods. FD stands for ``Feedback-Driven''. For HierFlow, ``coupled'' denotes bidirectional topology--execution updates rather than a fixed pipeline over two spaces.}
    \label{tab:comparison}
    \footnotesize
    \centering
    
    \setlength\tabcolsep{3.5pt}
    
    \begin{tabular}{lccccc}
        \toprule
        \toprule

        Method & Scope & FD & Search Space & Primary Cost\\

        \midrule

        ADAS & Task-Level & \textcolor{red}{\ding{55}} & Code & Offline Search \\

        AutoFlow & Task-Level & \textcolor{red}{\ding{55}} & NL Program & Offline Search \\

        AFlow & Task-Level & \textcolor{red}{\ding{55}} & Code & Offline Search \\

        MaAS & Query-Level & \textcolor{red}{\ding{55}} & Handcrafted & Offline Training \\

        DyFlow & Query-Level & \textcolor{green}{\ding{51}} & Handcrafted & Offline Training \\

        Flow & Query-Level & \textcolor{green}{\ding{51}} & Topology & Orchestration \\

        \midrule

        \sysname{} & \multirow{2}{*}{Query-Level} & \multirow{2}{*}{\textcolor{green}{\ding{51}}} & Coupled \& & Test-Time \\

        (Ours)& & & Topology--Code & Scaling \\

        \bottomrule
        \bottomrule
    \end{tabular}
    
\end{wraptable}

\noindent \textbf{Automatic Agentic Workflow Generation.} Recent methods automate workflow generation by formulating it as a search-based optimization problem over a proxy space. Based on the scope of generated workflow, they can be categorised into task-level (\textit{i.e.} dataset-level) and query-level approaches. Task-level methods aim to discover a general workflow for a given task or dataset, typically through offline search in code (\textit{e.g.,} ADAS~\cite{adas} and AFlow~\cite{aflow}) or NL program spaces (\textit{e.g.,} AutoFlow~\cite{autoflow}). In contrast, query-level methods generate workflows tailored to individual queries to improve generalisation. Representative examples include MaAS~\cite{maas}, which learns to select from predefined architectures, and feedback-driven approaches such as DyFlow~\cite{dyflow} and Flow~\cite{flow}, which dynamically adjust workflows during execution.

Despite their effectiveness, existing methods remain limited in terms of generalization and training cost. Task-level methods are constrained by the training domain distribution and often fail to generalize to unseen domains. Among query-level methods, AFlow and MaAS require pre-constructed training datasets, leading to high training costs and similarly limiting their generalization. Flow avoids training and is thus task-agnostic, but does not explicitly address execution-level correctness. Moreover, these methods typically operate within a single proxy space, overlooking the complementary strengths of different spaces and the potential benefits of jointly leveraging them. 
We propose \sysname{} to address these limitations, with a comparison summarized in Table~\ref{tab:comparison}.


\section{Methodology}
\label{sec:method}
\begin{figure*}[t]
    \centering
    \includegraphics[width=\linewidth]{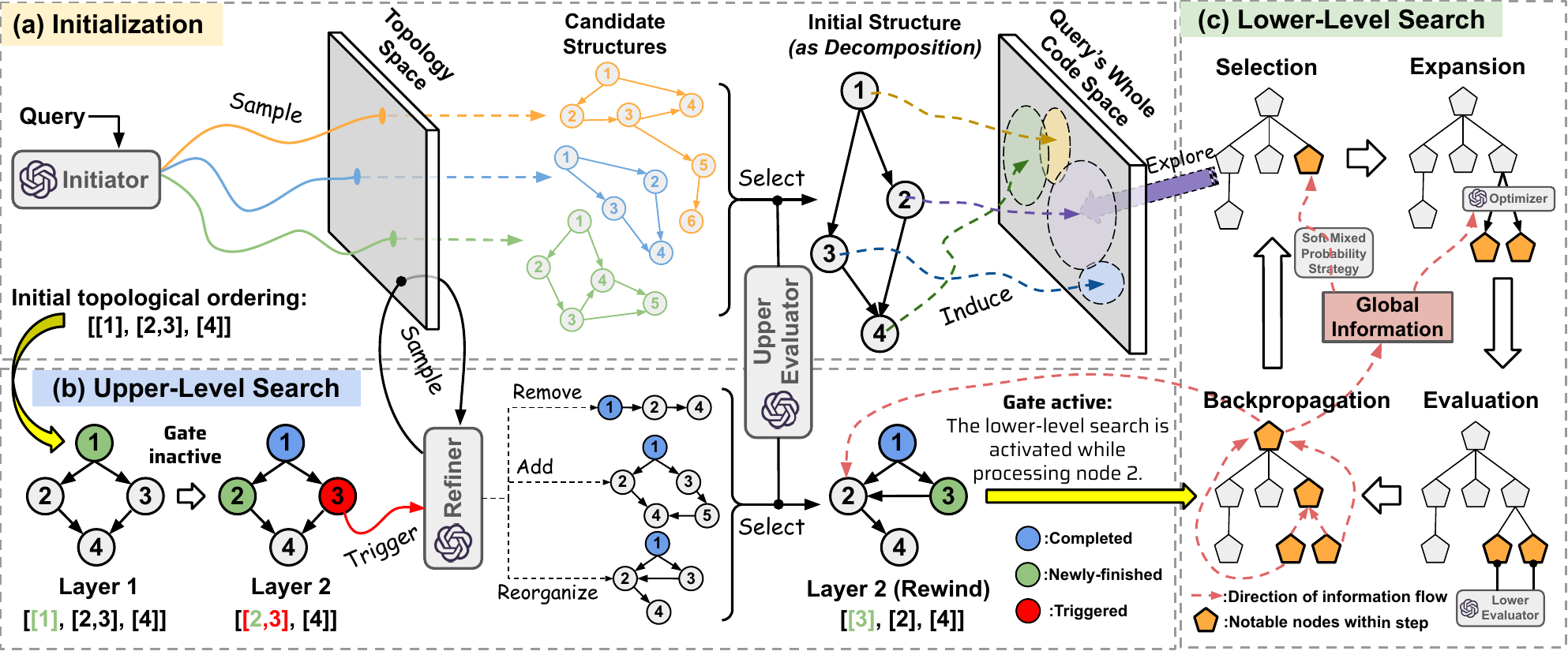}
    \caption{
    Illustrative example of how \sysname{} operates at inference time.
Given a query, \sysname{} first performs \textbf{(a) initialization} by identifying an initial structure in the topology space, where each node represents a subtask and corresponds to a sub-space of the overall workflow space. Based on this structure, (\textbf{b) upper-level search} is conducted to evaluate whether the current structure requires refinement, with a gating mechanism jointly determining whether any subtask should be activated for \textbf{(c) lower-level search}, which applies an MCTS-inspired lightweight tree search to explore a sub-workflow in the corresponding code space. The key distinction is the closed-loop coupling: topology determines the subtask-level code-search domains, while execution feedback can trigger topology repair and rewinding. See Section~\ref{sec:method} for more details.
    }
    \label{fig:overall}
\end{figure*}

To address the combinatorial complexity of automatic agentic workflow generation at test time, we propose \sysname{}, a hierarchical search framework that operates over a coupled dual space consisting of an upper-level task-topology space and a lower-level code execution space.
As illustrated in Figure~\ref{fig:overall}, \sysname{} comprises three components:
(i) an upper-level search that refines task decompositions in the topology space,
(ii) a MCTS-inspired lightweight tree search at the lower level, and
(iii) a gating mechanism for adaptive hierarchical control.
In addition, we provide a coupling-aware analysis that characterizes when hierarchical decomposition and gating are expected to be beneficial, and when their advantages may degrade under stronger cross-subtask coupling.
Further algorithmic details are provided in Appendix~\ref{app:algo_details}.

\subsection{Hierarchical  Search on Dual Space}
\label{subsec:dual_space_search}

Given a query $q\in\mathcal{Q}$ and a space of workflows $\mathcal{W}$, where each $w \in \mathcal{W}$ represents a complete agentic workflow. Automatic agentic workflow generation is typically formulated as the following optimization problem:
\begin{equation*}
    \begin{aligned}
    w^*(q) = \mathop{\text{arg max}}_{w \in \mathcal{W}}\ \mathcal{J}(w,q),
    \end{aligned}
\end{equation*}
where $\mathcal{J}:\mathcal{W}\times\mathcal{Q}\rightarrow\mathbb{R}$ is an evaluator for $w$.
In practice, a proxy space $\mathcal{S}$ is often adopted to replace $\mathcal{W}$ to facilitate search, as discussed in Section~\ref{sec:related}.


Existing methods commonly suffer from limited generalization capability and high training costs, whereas test-time scaling, as an inference-time paradigm with limited budget, can effectively address these issues. However, directly performing test-time search over $\mathcal{S}$ without prior knowledge remains both inefficient and brittle. We observe that different proxy spaces exhibit complementary strengths with significant potential for exploitation. 
Inspired by the principle of divide-and-conquer, we therefore construct a hierarchical dual-space proxy $\mathcal{S}$ to perform hierarchical search. Formally, HierFlow searches over a coupled state
\[
\Xi^{(T)}=\big(G^{(T)}, \{w_i^{(T)}\}_{i=1}^{N_T}\big),
\]
where $G^{(T)}=(V^{(T)},E^{(T)})$ is the topology at iteration $T$, $N_T=|V^{(T)}|$, and $w_i^{(T)}$ is the current executable sub-workflow for node $v_i^{(T)}$. Both parts of $\Xi^{(T)}$ are search variables updated through interaction:
\[
w_i^{(T)}
\in
\mathcal{W}_i\!\left(G^{(T)}, v_i^{(T)}, \mathrm{Pa}_{G^{(T)}}(v_i^{(T)})\right),
\qquad
G^{(T+1)}
=
U\!\left(G^{(T)}, F^{(T)}(\{w_i^{(T)}\}_{i=1}^{N_T})\right),
\]
where $F^{(T)}$ aggregates lower-level execution feedback. Thus, the topology defines the code-search domains at iteration $T$, and the resulting executions define the topology revision at iteration $T{+}1$. In contrast, a fixed-topology pipeline freezes $G=G^{(0)}$ and reduces the problem to optimizing only $\{w_i\}$ under a committed structure.

Specifically, at the upper-level task-topology space, we search a task-decomposition space to construct modular, dependency-aware workflow graphs that capture reusable problem-solving structures, while at the lower-level code execution space, we perform targeted test-time search over executable code workflows for individual subtasks. 
Consistent with prior work on LLM-based agentic reasoning~\cite{CoT} and modularized workflows~\cite{flow}, we use a DAG-based task decomposition to organize the search. However, we do not require the subtasks to be perfectly independent. Instead, the topology is used as a controllable proxy that localizes execution-level search, while residual cross-subtask coupling is explicitly tracked through an observable coupling measure.

\begin{definition}[Observable Coupling Proxy]
\label{def:coupling-proxy}
Given a task decomposition $G=(V^t,E^t)$ with $N=|V^t|$, let $d_i=\deg(v_i^t)$ and $\bar d=\frac{1}{N}\sum_i d_i$. We define the observable coupling proxy as
\begin{equation}
\widehat{\kappa}(G)
=
\underbrace{\frac{|E^t|}{N}}_{\text{edge density}}
+
\beta
\underbrace{
\sqrt{\frac{1}{N}\sum_{i=1}^{N}(d_i-\bar d)^2}
}_{\text{degree imbalance}}
+
\eta
\underbrace{
\frac{1}{N}\sum_{i=1}^{N} o_i^t
}_{\text{execution-observed repair rate}},
\label{eq:coupling-proxy}
\end{equation}
where $o_i^t$ is the refinement-triggering signal returned by the lower level in Eq.~(3). Before execution feedback is available, the last term is set to zero. The constants $\beta,\eta\ge 0$ control the relative weights of structural and execution-observed coupling.
\end{definition}

Definition~\ref{def:coupling-proxy} turns the original weak-coupling condition into a measurable quantity. A small $\widehat{\kappa}(G)$ indicates that the decomposition has sparse and balanced dependencies with limited repair feedback, while a large $\widehat{\kappa}(G)$ indicates stronger cross-subtask interaction, higher dependency concentration, or more execution-triggered topology repair.

\begin{theorem}[Coupling-Aware Cost Accounting]
\label{thm:coupling-cost}
For a realized \sysname{} run with topology $G=(V^t,E^t)$, lower-level search costs $\{c_i\}_{i=1}^{N}$, and gating decisions $\{o_i^g\}_{i=1}^{N}$, let
\[
\kappa_R(G)=\frac{1}{N}\sum_{i=1}^{N}o_i^t
\]
denote the execution-observed repair-rate component of $\widehat{\kappa}(G)$. The test-time cost can be decomposed as
\[
C_{\mathrm{hier}}
=
C_{\mathrm{hier}}^{\mathrm{upper}}(N,|E^t|)
+
\sum_{i=1}^{N}o_i^g c_i
+
C_{\mathrm{repair}}(\kappa_R),
\]
where
\[
C_{\mathrm{hier}}^{\mathrm{upper}}(N,|E^t|)
=
O(c_{\max}^t K (N+|E^t|))
\]
is the topology-level coordination cost under refinement budget $c_{\max}^t$ and $K$ candidate decompositions per refinement step, and $C_{\mathrm{repair}}(\kappa_R)$ captures topology repair, rewinding, synchronization, and re-execution induced by lower-level feedback. Thus, the edge-density component of $\widehat{\kappa}(G)$ affects the upper-level coordination term through $|E^t|$, while the execution-observed repair component $\kappa_R$ affects the repair term. Consequently, hierarchical search has the clearest efficiency advantage when observable coupling is small or moderate, while its advantage may degrade as dependency density or repair burden increases.
\end{theorem}

\begin{remark}
\label{rem:theory-scope}
Theorem~\ref{thm:coupling-cost} is a coupling-aware characterization rather than an unconditional asymptotic speedup guarantee. It explains the intended operating regime of \sysname{}: topology-level decomposition is most useful when it localizes execution search, but denser dependencies, strong global interactions, or repeated repair feedback can reduce this benefit. The proof and additional discussion are provided in Appendix~\ref{app:coupling-cost}.
\end{remark}




\subsection{Upper-Level Topology Search via Refinement}
\label{sec:upper-level-search}

The goal of the upper-level search is to efficiently orchestrate a \emph{reasonable} task decomposition for a given query $q \in \mathcal{Q}$. Rather than exhaustively enumerating workflow structures, the upper level focuses on identifying a structurally sound and execution-feasible decomposition that can be progressively refined through feedback from lower-level.

\noindent \textbf{Topology Evaluation.}
We first establish an evaluation criterion for a topology $G=(V,E)$. Specifically, we define:
\begin{equation}
J(G) \;=\; \lambda_0\,J_{\mathrm{dep}}(G)\;+\;\sum_{m\in\{\mathrm{cov},\mathrm{snd},\mathrm{sea}\}}
\lambda_m\,J_m(G),
\label{eq:upper-level-eval}
\end{equation}
where $J_{\mathrm{dep}}(G)$ is a dependency-complexity term from Flow~\cite{flow}.
The remaining terms $J_{\mathrm{cov}}(G)$, $J_{\mathrm{snd}}(G)$, and $J_{\mathrm{sea}}(G)$ are scored by a LLM-based evaluator, corresponding to task coverage and decomposition sufficiency, structural soundness and dependency consistency, and structural complexity and searchability,
respectively.


\noindent \textbf{Initialization.} Given a query $q$, an initiator generates candidate workflow structures $\{G_k\}_{k=1}^K$, from which the initial structure $G^{(0)}$ is selected by maximizing $J(G_k)$. A layer-wise topological ordering $\{\mathcal{L}_\ell\}_{\ell=1}^L$ is then computed via a batched variant of Kahn’s algorithm~\cite{kahn}.

\noindent \textbf{Layer-Wise Execution.} The upper-level search traverses the decomposition layer by layer: for each layer $\mathcal{L}_\ell$, all subtasks $v^t_i \in \mathcal{L}_\ell$ are processed in parallel, and a lower-level search is invoked to produce a candidate sub-workflow $w_i$ along with feedback. 

\noindent \textbf{Topology Refinement.} If any subtask $v_i^t$ receives a refinement-triggering signal $o^t_i$ (Eq.\eqref{eq:refinement-trigger}) and the corresponding failure feedback $\mathcal{F}^{(T)}$ from the lower level, and the refinement budget $c_{\max}^t$ has not been exhausted, the task decomposition is incrementally updated as
$G^{(T+1)} = \mathcal{U}\!\left(G^{(T)}, \mathcal{F}^{(T)}\right)$,
where $\mathcal{U}$ is a topology refiner which generates a revised decomposition $G^{(T+1)}$ by modifying, adding or removing subtasks, and reorganizing dependencies on $G^{(T)}$, while preserving all successfully completed subtasks.
In practice, $\mathcal{U}$ is instantiated by a LLM, which proposes $K$ candidate decompositions and selects the highest-scoring one according to Eq.\eqref{eq:upper-level-eval}. After the update, the layer-wise ordering is recomputed and traversal resumes from the earliest unfinished subtask.

\noindent \textbf{Termination.} The procedure terminates once all subtasks are finished, yielding a complete workflow composed of the refined structure and the corresponding subtask implementations. Additionally, we maintain a memory bank that stores key--value pairs of subtask descriptions and their corresponding lower-level configurations, enabling efficient reuse when semantically similar subtasks are encountered. 

\noindent \textbf{Complexity Analysis.} The upper-level topology process has coordination cost $O(c_{\max}^t K (N+|E^t|))$, where $K$ is the number of candidate decompositions and $c_{\max}^t$ is the refinement budget. When the induced topology has moderate observable coupling, i.e., sparse dependencies and limited execution-triggered repair, this cost is close to linear in the number of subtasks. Together with gating, which limits the number of activated lower-level searches, this yields a budgeted hierarchical search procedure. When observable coupling increases, the repair and re-execution terms may grow, which is the main boundary of the theoretical efficiency argument. See Appendix~\ref{app:coupling-cost} for details.



\subsection{MCTS-inspired Lower-Level Code Search}
\label{sec:lower-level-search}

The lower-level search aims to generate high-quality sub-workflow $w_i$ implementations for subtasks $v^t_i$ provided by the upper level. Through test-time scaling, the lower level adaptively allocates computation to explore and refine execution strategies, producing execution outcomes that inform upper-level structural updates.
Compared to the overall task, each subtask is defined over a narrower interface and fewer degrees of freedom, resulting in a significantly smaller and more structured execution search space. 

Accordingly, we adopt a MCTS-inspired lightweight tree search to explore executable code within individual subtasks.
In lower-level search tree with maximum depth $d_{\max}$ and branching factor $b_{\max}$, each node represents a complete candidate sub-workflow, and each parent--child edge corresponds to a causal modification that transforms the parent sub-workflow into the child. When a subtask $v_i^t$ is newly created at the upper level, its lower-level search tree $\mathcal{T}_i$ is initialized with a root node $v_1^c$ produced by a direct input--output (IO) generation attempt to solve the subtask. The search process builds $T_i$ by iterating over four MCTS-inspired phases:

\noindent \textbf{Selection.} For \textbf{all} previously explored nodes $\{v_j^c\}_{j=1}^{M_i}$ in $\mathcal{T}_i$, sampling is performed according to the soft mixed probability strategy from AFlow~\cite{aflow}.

\noindent \textbf{Expansion.} An LLM-based optimizer expands the selected node to obtain a new node $v_{j'}^c$ by applying operators widely adopted in prior work~\cite{maas,dyflow} based on the node’s contextual information. Unlike conventional MCTS, our nodes inherit execution history without backpropagating scores to parents. 
This decoupling enables one-step parallel expansion, allowing the optimizer to generate and explore multiple diverse candidate strategies simultaneously. 

\noindent \textbf{Evaluation.} We evaluate newly generated sub-workflows using an execution-grounded scheme, in which each candidate first undergoes a hard executability check with schema-driven mock inputs executed in a sandbox; failures yield a failure signal $o^e_{j'}=1$. Executable workflows are then assigned a unified quality score $s_{j'}^c$ and corresponding evaluation rationale $r_{j'}^c$ by an LLM-based evaluator $J_{sub}$ with execution feedback, reflecting semantic completeness, interface compliance, feasibility, and output consistency.

\noindent \textbf{Backpropagation.} The evaluation signals of newly expanded nodes are propagated back to the root node, while the corresponding modification operations relative to the parent, together with their optimization outcomes relative to the parent node, are propagated back to the parent.

\noindent \textbf{Termination.} The lower-level search terminates when the budget $c_{max}^c$ exhausted, the search tree reaches its maximum size, or the optimizer determines that further expansion is no longer beneficial. Upon termination, the highest-scoring node is selected as the baseline sub-workflow for $v_i^t$, and a refinement signal $o_i^t$, which indicates whether the current subtask requires upper-level refinement, is propagated to the upper level and computed as follows:
\begin{equation}
o_i^t
=
\mathbb{I}\!\left[
\sum_{j=1}^{M_i} o_j^e \ge \frac{M_i}{2}
\;\lor\;
\mathrm{Var}\!\left(s_j^c \mid o_j^e = 0\right) > \gamma
\right],
\label{eq:refinement-trigger}
\end{equation}
where $\mathbb{I}\{\cdot\}$ denotes the indicator function, and $\gamma>0$ is a threshold controlling the sensitivity of refinement.

\subsection{Adaptive Hierarchical Control via Gating}
\label{sec:adaptive-control}

A central challenge in hierarchical search is deciding \emph{when} to invoke lower-level search.
Blindly triggering execution-level search for every subtask is computationally wasteful, while deferring execution feedback risks propagating infeasible decompositions.
To address this trade-off, we introduce a test-time gating mechanism for adaptive hierarchical control, which selectively mediates the interaction between upper-level task decomposition and lower-level execution search for effective computational resource allocation.

\noindent \textbf{Gating Mechanism.} Specifically, for each subtask $v_i^t \in V^t$, the gating decision is informed by two complementary signals: the \emph{search potential} and the \emph{uncertainty} of the current solution. The search potential $S_i$ quantifies the \emph{unexplored} workflow space for $v_i^t$, defined as the logarithmic difference between the total candidate space $\mathcal{W}_i$ and the portion explored $\mathcal{T}_i$ so far,
\begin{equation*}
S_i = \log\!\big(|\mathcal{W}_i| - |\mathcal{T}_i|\big)
\;\approx\;
\log\!\left(b_{\max}^{d_{\max}} - M_i\right).
\end{equation*}
In parallel, the uncertainty $U_i$ captures the robustness of the baseline workflow for $v_i^t$. Specifically, we estimate uncertainty using the predictive entropy~\cite{predictive} of its evaluation rationale $r_i^t$,
\begin{equation*}
U_i
=
\mathbb{E}_{r_i^t}\!\left[
\frac{1}{|r_i^t|}
\sum_{a_j \in r_i^t}
-\log \mathbb{P}\!\left(a_j \mid a_0, \ldots, a_{j-1}\right)
\right],
\end{equation*}
where $a_j$ denotes the $j$-th token in the rationale sequence.
Intuitively, larger values of $S_i$ and $U_i$ respectively indicate substantial remaining search space and insufficient robustness of the baseline, both suggesting that further exploration may be beneficial.
Motivated by this intuition, we combine these two signals into a single proxy that estimates the necessity of invoking lower-level search.

\begin{definition}[Search Necessity Proxy]
\label{def:search-necessity-proxy}
Given the search potential $S_i$ and uncertainty $U_i$ of subtask $v_i^t$, the search necessity proxy is defined as $ Z_i \;\triangleq\; S_i \cdot U_i .$
\end{definition}

Based on Definition~\ref{def:search-necessity-proxy}, the \emph{gating mechanism} activates lower-level search for subtask $v_i^t$ according to $o_i^g = \mathbb{I}\{Z_i > \tau\},$
where $\tau>0$ is a threshold controlling the activation rate: smaller values encourage more aggressive search at higher computational cost, while larger values enforce more conservative gating. Rather than assuming that $Z_i$ is universally faithful to the true gain of search, we interpret gating as a proxy-based ranking rule whose usefulness can be empirically measured. Let $\Delta Q_i$ denote the realized quality gain obtained by invoking lower-level search for subtask $v_i^t$ under a fixed lower-level budget. We measure the informativeness of the gating proxy by the rank correlation between $Z_i$ and $\Delta Q_i$.

\begin{definition}[Proxy Informativeness]
\label{def:proxy-informativeness}
Given a set of subtasks, the empirical informativeness of the gating proxy is
\[
\varrho_Z
=
\operatorname{Spearman}
\left(
\{Z_i\}_{i=1}^{N},
\{\Delta Q_i\}_{i=1}^{N}
\right),
\]
where larger $\varrho_Z$ indicates that the proxy ranking is more aligned with the realized lower-level search gains.
\end{definition}

The test-time gating problem can still be viewed as a budgeted allocation problem over subtasks:
\begin{equation}
\max_{\{o_i^g\}_{i=1}^{N}}
\mathbb{E}
\left[
\sum_{i=1}^{N} Q_i(o_i^g)
\right]
\quad
\text{s.t.}
\quad
C_{\mathrm{hier}}^{\mathrm{upper}}
+
\sum_{i=1}^{N} o_i^g c_i
\le B,
\tag{4}
\end{equation}
where $Q_i(1)$ and $Q_i(0)$ denote the solution quality obtained with and without lower-level search, respectively, and $c_i$ is the cost of searching subtask $i$.

\begin{theorem}[Rank-Based Interpretation of Gating]
\label{thm:gating-rank}
Under approximately uniform lower-level search costs, a budget that activates $k$ subtasks reduces Eq.~(4) to a top-$k$ selection problem. Let $\mathcal{O}_k$ be the oracle top-$k$ set ranked by the realized gains $\Delta Q_i$, and let $\mathcal{P}_k$ be the proxy top-$k$ set ranked by $Z_i$. If $\epsilon_k = |\mathcal{O}_k \triangle \mathcal{P}_k|/(2k)$ denotes their top-$k$ disagreement rate and $\Delta Q_i\in[\Delta_{\min},\Delta_{\max}]$, then
\[
\sum_{i\in\mathcal{O}_k}\Delta Q_i
-
\sum_{i\in\mathcal{P}_k}\Delta Q_i
\le
k\epsilon_k(\Delta_{\max}-\Delta_{\min}).
\]
Thus, the gap between threshold gating and oracle gating is controlled by their top-$k$ rank disagreement. A positive $\varrho_Z$ and increasing realized gains across higher-$Z_i$ bins provide empirical evidence that the gating proxy is informative for budget allocation.
\end{theorem}

\begin{remark}
\label{rem:gating-scope}
Theorem~\ref{thm:gating-rank} does not claim unconditional near-optimality of $Z_i=S_iU_i$. Instead, it makes the required condition explicit: the proxy should rank high-gain subtasks above low-gain subtasks more often than chance. This condition is directly testable through $\varrho_Z$ and improvement-rate stratification. The proof and additional discussion are provided in Appendix~\ref{app:gating-rank}.
\end{remark}

\section{Experiments}
\label{sec:exp}
To evaluate the effectiveness of \sysname{}, we conduct comprehensive experiments across five benchmark datasets, comparing against eleven workflow generation baselines. Our study is guided by the following research questions: \textbf{(RQ1)} Can \sysname{} discover higher-quality workflows than both manual and automatic baselines without relying on additional training data? \textbf{(RQ2)} What is the test-time inference cost of \sysname{}, and how does it trade off against workflow quality under different gating thresholds $\tau$? \textbf{(RQ3)} How does \sysname{} generalize to new data distributions compared to existing automatic baselines? \textbf{(RQ4)} How do the key components of \sysname{}, including dual-space proxy, hierarchical search, and the gating mechanism, individually and jointly affect performance and computational cost?

\newcommand{\scorestd}[2]{#1\raisebox{-0.45ex}{\scriptsize\ensuremath{\pm}#2}}
\newcommand{\bscorestd}[2]{\textbf{#1\raisebox{-0.45ex}{\scriptsize\ensuremath{\pm}#2}}}
\newcommand{\uscorestd}[2]{\underline{#1\raisebox{-0.45ex}{\scriptsize\ensuremath{\pm}#2}}}

\subsection{Experimental Setup} 

\noindent \textbf{Benchmarks.}
We conduct evaluation on diverse benchmarks covering: \textbf{(i) question answering (QA)} using \textsc{HotpotQA}~\cite{HotpotQA} evaluated by F1 score, \textbf{(ii) mathematical reasoning} using \textsc{GSM8K}~\cite{GSM8K} and \textsc{MATH}~\cite{MATH} evaluated by accuracy, and (\textbf{iii) code generation} using \textsc{HumanEval}~\cite{HumanEval} and \textsc{MBPP}~\cite{MBPP} evaluated by pass@1.
Additional details are provided in Appendix~\ref{app:datasets}.


\noindent \textbf{Baselines.} We compare against two categories of baselines: \textbf{(i) manually designed methods}, including IO, CoT~\cite{CoT}, Self-Consistency with five CoT paths (SC@5)~\cite{CoTSC}, Self-Refine with three iterations~\cite{SR}, and LLM-Debate~\cite{llmdebate}; and \textbf{(ii) automatic workflow generation methods}, including ADAS~\cite{adas}, AutoFlow~\cite{autoflow}, AFlow~\cite{aflow}, Flow~\cite{flow}, MaAS~\cite{maas}, and DyFlow~\cite{dyflow}. 
All baselines are implemented using the optimal settings reported in their original papers. See Appendix~\ref{app:baseline} for more baseline implementation details.

\noindent \textbf{Implementation Details.} All LLM oracles in \sysname{} are instantiated with gpt-4o-mini to ensure alignment with the baselines.
In the upper-level search, the weights in Eq.~\eqref{eq:upper-level-eval} are set uniformly, the number of candidate topologies is fixed to $K=3$, and the refinement budget is set to $c_{\max}^t = 20$.
For the lower-level search, the search tree depth and branching factor are constrained to $d_{\max} = 3$ and $b_{\max} = 3$, respectively, with a per-search budget of $c_{\max}^c = 5$.
The gating threshold $\tau$ is set to 4, balancing solution quality and test-time computational cost. Results average three independent runs. 
A discussion of the role of LLM-based evaluation in HierFlow, together with judge-model sensitivity and evaluator–end-task agreement analyses, is provided in Appendix~\ref{appendix:judge_robustness}.

\begin{table*}[t]
\centering
\footnotesize
\caption{Overall performance of manual and automatic workflow generation methods across different scenarios. The lower-right subscript reports the standard deviation. The best results are highlighted in \textbf{bold}, and the runner-up results are \underline{underlined}.}
\begin{tabular}{l|cccccc}
\toprule
 & \textsc{HotpotQA} & \textsc{GSM8K} & \textsc{MATH} & \textsc{MBPP} & \textsc{HumanEval} & \textsc{Avg.} \\ 
\midrule
Vanilla (GPT-4o-mini) 
& \scorestd{67.25}{0.42} 
& \scorestd{87.39}{0.31} 
& \scorestd{47.33}{0.76} 
& \scorestd{71.85}{0.68} 
& \scorestd{87.02}{1.14} 
& 72.17 \\ 
\cmidrule(lr){1-7}

CoT~\cite{CoT} 
& \scorestd{67.63}{0.55} 
& \scorestd{87.49}{0.37} 
& \scorestd{48.15}{0.89} 
& \scorestd{71.85}{0.74} 
& \scorestd{88.55}{0.96} 
& 72.73 \\

SC@5~\cite{CoTSC} 
& \scorestd{68.50}{0.63} 
& \scorestd{87.87}{0.44} 
& \scorestd{49.18}{1.05} 
& \scorestd{73.61}{0.91} 
& \scorestd{89.31}{1.22} 
& 73.69 \\

Self-Refine~\cite{SR} 
& \scorestd{66.00}{0.71} 
& \scorestd{85.40}{0.52} 
& \scorestd{46.71}{0.94} 
& \scorestd{69.79}{1.08} 
& \scorestd{87.79}{1.37} 
& 71.14 \\

LLM-Debate~\cite{llmdebate} 
& \scorestd{67.25}{0.59} 
& \scorestd{87.87}{0.48} 
& \scorestd{50.00}{1.18} 
& \scorestd{70.38}{0.83} 
& \scorestd{88.55}{1.05} 
& 72.81 \\

\cmidrule(lr){1-7}

ADAS~\cite{adas} 
& \scorestd{66.63}{0.88} 
& \scorestd{86.35}{0.61} 
& \scorestd{42.80}{1.27} 
& \scorestd{68.04}{1.19} 
& \scorestd{84.73}{1.46} 
& 69.71 \\

AutoFlow~\cite{autoflow} 
& \scorestd{67.50}{0.76} 
& \scorestd{89.19}{0.57} 
& \scorestd{50.41}{1.11} 
& \scorestd{77.71}{0.96} 
& \scorestd{88.55}{1.31} 
& 74.67 \\

AFlow~\cite{aflow} 
& \uscorestd{70.13}{0.64} 
& \scorestd{91.18}{0.43} 
& \scorestd{51.44}{0.97} 
& \scorestd{81.23}{0.72} 
& \scorestd{90.84}{1.08} 
& 76.96 \\

MaAS~\cite{maas} 
& \scorestd{69.00}{0.69} 
& \uscorestd{92.51}{0.39} 
& \scorestd{51.85}{1.04} 
& \uscorestd{82.11}{0.84} 
& \scorestd{91.60}{0.93} 
& \underline{77.42} \\

DyFlow~\cite{dyflow} 
& \scorestd{67.88}{0.81} 
& \scorestd{91.47}{0.46} 
& \uscorestd{52.67}{0.86} 
& \scorestd{78.89}{1.02} 
& \uscorestd{92.37}{0.79} 
& 76.65 \\

Flow~\cite{flow} 
& \scorestd{67.63}{0.73} 
& \scorestd{88.53}{0.54} 
& \scorestd{48.15}{1.13} 
& \scorestd{77.71}{0.89} 
& \scorestd{89.31}{1.16} 
& 74.27 \\

\cmidrule(lr){1-7}

\textbf{\sysname{} } 
& \bscorestd{71.75}{0.58} 
& \bscorestd{92.70}{0.35} 
& \bscorestd{53.70}{0.91} 
& \bscorestd{82.40}{0.77} 
& \bscorestd{93.13}{0.88} 
& \textbf{78.74} \\

\bottomrule
\end{tabular}
\vspace{-4mm}
\label{tab:main-result}
\end{table*}

\subsection{Overall Performance (RQ1)}
Overall performance are reported in Table~\ref{tab:main-result}. \sysname{} consistently achieves the best performance across all datasets without using any training data, outperforming both manually designed and automatically generated methods. This result indicates that \sysname{} produces higher-quality workflows, reflecting strong usability and robustness across tasks. 
While both MaAS and DyFlow demonstrate competitive performance, they rely on offline data construction, which incurs additional cost. In particular, MaAS requires training data for each new data distribution, whereas DyFlow relies on an offline-trained designer without requiring dataset reconstruction for every benchmark. On MATH, a dataset used to train its designer, DyFlow achieves the highest baseline mean (52.67\%), while HierFlow obtains a higher observed mean of 53.70\% without dataset-specific offline optimization.
Although Flow also relies on test-time refinement, its focus on orchestration causes the performance of the generated workflows to depend heavily on the quality of task decomposition. 
In contrast, the lower-level search in \sysname{} provides a safeguard for workflow quality and, 
through richer feedback, further facilitates more effective task decomposition at the upper level.

\subsection{Efficiency–Quality Trade-off Analysis (RQ2)}
\label{subsec:cost_analysis}

\begin{wrapfigure}{R}{0.6\textwidth}
    \centering
    \includegraphics[width=\linewidth]{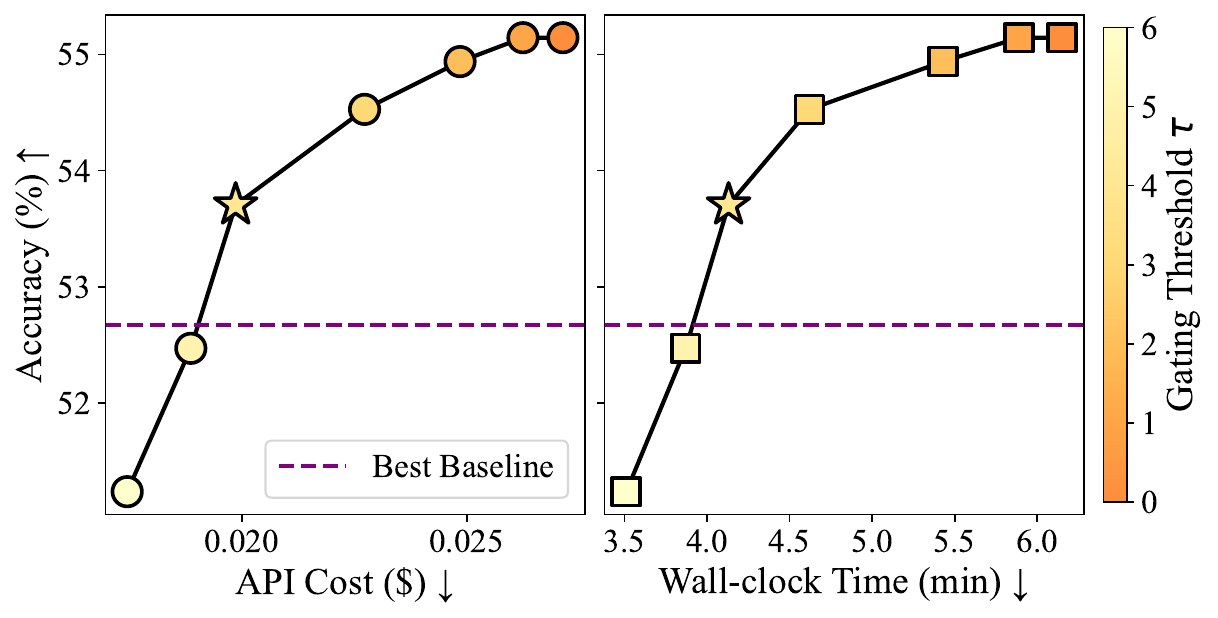}
    \caption{
    Pareto fronts illustrating the trade-off between accuracy and efficiency of \sysname{} on the \textsc{MATH} dataset per query under different gating thresholds $\tau$, where higher accuracy ($\uparrow$) and lower cost or wall-clock time ($\downarrow$) are preferred. Points closer to the upper-left corner indicate more favorable accuracy--efficiency trade-offs. The star marker highlights the selected operating point at $\tau=4$, while the dashed horizontal line denotes the best baseline accuracy.
    }
    \label{fig:cost_analysis}
\end{wrapfigure}
We study the quality--efficiency trade-off of \sysname{} by varying the gating threshold $\tau$ from 0 to 6, which controls how often lower-level search is activated. As shown in Figure~\ref{fig:cost_analysis}, larger $\tau$ values make gating more conservative, reducing API cost and wall-clock time but gradually lowering accuracy; smaller $\tau$ values trigger more aggressive search and improve accuracy at higher cost. This smooth Pareto trend shows that $\tau$ provides an interpretable control knob for test-time scaling and empirically supports the rank-based gating interpretation in Theorem~\ref{thm:gating-rank}. The selected operating point, $\tau=4$, lies near the knee of the curve, achieving strong accuracy while substantially reducing cost compared with more aggressive configurations. Further analyses of gating-proxy informativeness and the inference-time cost--performance trade-off with Flow are provided in Appendices~\ref{appendix:gating_proxy} and~\ref{app:flow_cost}, respectively.

\subsection{Generalization Analysis (RQ3)}

To evaluate deployment robustness across data distributions, we consider three transfer settings: \textsc{MATH}$\rightarrow$\textsc{GSM8K} for mathematical reasoning, \textsc{HumanEval}$\rightarrow$\textsc{MBPP} for code generation, and \textsc{HumanEval}$\rightarrow$\textsc{HotpotQA} for cross-task transfer. As shown in Table~\ref{tab:generalization}, methods that depend on task-specific training or offline search, including ADAS, AutoFlow, AFlow, and MaAS, suffer clear performance degradation under distribution shifts, especially in the cross-task setting. 
\begin{wraptable}{r}{0.6\textwidth}
\centering
\footnotesize
\setlength{\abovecaptionskip}{-1pt}
\caption{Performance of automatic workflow generation methods under three transfer settings. The \textcolor{purple}{$\downarrow$} indicates performance degradation relative to the in-domain results.}
\begin{tabular}{l|ccc}
\toprule
\multirow{2}{*}{
  \makecell[c]{\textit{Training}\\[-3pt]$\downarrow$\\[-3pt]\textit{Testing}}
} & \textsc{MATH} & \multicolumn{2}{c}{\textsc{HumanEval}} \\
\cmidrule(lr){2-4}
 & \textsc{GSM8K} & \textsc{MBPP}  & \textsc{HotpotQA} \\
\midrule
ADAS     & 84.45$_{\scriptstyle\,\textcolor{purple}{\downarrow1.90}}$
         & 65.40$_{\scriptstyle\,\textcolor{purple}{\downarrow2.64}}$
         & 62.75$_{\scriptstyle\,\textcolor{purple}{\downarrow3.88}}$ \\

AutoFlow & 87.20$_{\scriptstyle\,\textcolor{purple}{\downarrow1.99}}$
         & 74.78$_{\scriptstyle\,\textcolor{purple}{\downarrow2.93}}$
         & 63.50$_{\scriptstyle\,\textcolor{purple}{\downarrow4.00}}$ \\

AFlow    & 89.38$_{\scriptstyle\,\textcolor{purple}{\downarrow1.80}}$
         & 78.89$_{\scriptstyle\,\textcolor{purple}{\downarrow2.35}}$
         & 66.63$_{\scriptstyle\,\textcolor{purple}{\downarrow3.50}}$ \\

MaAS     & 90.81$_{\scriptstyle\,\textcolor{purple}{\downarrow1.71}}$
         & 79.47$_{\scriptstyle\,\textcolor{purple}{\downarrow2.64}}$
         & 66.38$_{\scriptstyle\,\textcolor{purple}{\downarrow2.63}}$ \\

DyFlow   & 91.47$_{\scriptstyle\,\textcolor{purple}{\downarrow0.00}}$
         & 78.89$_{\scriptstyle\,\textcolor{purple}{\downarrow0.00}}$
         & 67.88$_{\scriptstyle\,\textcolor{purple}{\downarrow0.00}}$ \\

Flow     & 88.53$_{\scriptstyle\,\textcolor{purple}{\downarrow0.00}}$
         & 77.71$_{\scriptstyle\,\textcolor{purple}{\downarrow0.00}}$
         & 67.63$_{\scriptstyle\,\textcolor{purple}{\downarrow0.00}}$ \\
\cmidrule{1-4}
\textbf{\sysname{}} & \textbf{92.70}$_{\scriptstyle\,\textcolor{purple}{\downarrow0.00}}$
         & \textbf{82.40}$_{\scriptstyle\,\textcolor{purple}{\downarrow0.00}}$
         & \textbf{71.75}$_{\scriptstyle\,\textcolor{purple}{\downarrow0.00}}$ \\

\bottomrule
\end{tabular}
\label{tab:generalization}

\end{wraptable}
In contrast, DyFlow, Flow, and \sysname{} maintain results consistent with Table~\ref{tab:main-result} because they do not require rebuilding a workflow generator for each target dataset. However, DyFlow still depends on the datasets used to pretrain its designer. These results therefore suggest that \sysname{} is plug-and-play across new evaluation distributions and reduces exposure to train--test mismatch without additional training, rather than relying on dataset-specific workflow optimization.

\subsection{Ablation Study (RQ4)}

\begin{wraptable}{r}{0.6\textwidth}
\centering
\footnotesize
\setlength{\abovecaptionskip}{-4pt}
\setlength\tabcolsep{3pt}
\caption{Ablation study of \sysname{} on the \textsc{MATH} dataset. All metrics except accuracy (Acc.) are reported as averages per query.}
\begin{tabular}{l|ccccc}
\toprule
 & Prompt & Completion & Time & Cost & Acc.  \\
 & Token & Token & (min) & (\$) &(\%) \\
\midrule
\textit{w/o} topology & 56{,}253 & 45{,}532 & 8.21 & 0.036 & 51.23 \\
\textit{w/o} upper & 26{,}956 & 15{,}672 & 3.15 & 0.013 & 50.62 \\
\textit{w/o} lower & 13{,}425 & 7{,}217 & 1.27 & 0.006 & 48.77 \\
\textit{w/o} gating & 50{,}783 & 32{,}530 & 6.15 & 0.027 & 55.14 \\
\cmidrule(lr){1-6}
\sysname{} & 37{,}256 & 23{,}793 & 4.13 & 0.020 & 53.70 \\
\bottomrule
\end{tabular}
\label{tab:ablation-result}
\vspace{-2mm}
\end{wraptable}
To analyze the contribution of individual components in \sysname{}, we perform an ablation study with four variants: \textbf{(i) \textit{w/o} topology space}, where the lower-level search operates directly in the code space, effectively reducing to a flat single-space search;
\textbf{(ii) \textit{w/o} upper-level refinement}, which freezes the LLM-initialized topology and performs lower-level code search without execution-to-topology feedback;
\textbf{(iii) \textit{w/o} lower-level search}, applying only upper-level topology refinement with each sub-workflow generated by a single IO pass, equivalent to $\tau \rightarrow \infty$;
\textbf{(iv) \textit{w/o} gating mechanism}, where the gating is removed by setting $\tau = 0$, such that every upper-level expansion always triggers lower-level search.
The corresponding results are reported in Table~\ref{tab:ablation-result}.
Our results support coupled dual-space optimization. Removing topology yields the highest cost and suboptimal accuracy, consistent with Theorem~\ref{thm:coupling-cost} that topology localizes lower-level search. Freezing the initial topology in w/o upper degrades performance, suggesting that execution feedback helps revise the decomposition, not only improve sub-workflows. Removing lower-level search shows that topology refinement alone is insufficient for execution quality. Finally, w/o gating achieves higher accuracy only by uniformly activating lower-level search at much larger cost, while \sysname{} provides a more favorable budget-aware accuracy--cost trade-off. Further analysis is provided in Appendix~\ref{app:detailed_ablation}.

\section{Conclusion}
\label{sec:conclusion}
We proposed \sysname{}, a training-free hierarchical search framework for agentic workflow generation that operates at test time. By formulating workflow optimization as coupled topology–execution search with an upper-level topology space and a lower-level code execution space, \sysname{} mitigates the combinatorial complexity of workflow search while leveraging complementary proxy spaces. Feedback-driven topology refinement, MCTS-inspired subtask-level tree search, and an adaptive gating mechanism jointly enable efficient, budget-aware test-time optimization. We further provided a coupling-aware analysis explaining when hierarchical decomposition and proxy-based gating are expected to be beneficial, and where their advantages may degrade under stronger coupling, and extensive experiments demonstrate that \sysname{} achieves strong performance and favorable efficiency--quality trade-offs without additional training.

\section*{Acknowledgment}

This work is supported by the National Science Foundation (NSF) \# 2515265.



\bibliography{refer}
\bibliographystyle{abbrv}


\newpage
\appendix

\section{Proofs and Additional Theoretical Discussions}
\label{app:theory}

This appendix provides the details for the coupling-aware analysis in Section~\ref{sec:method}. The goal is not to prove an unconditional asymptotic speedup under idealized independence assumptions, but to characterize when hierarchical decomposition and proxy-based gating are expected to be useful.

\subsection{Proof of Theorem~\ref{thm:coupling-cost}}
\label{app:coupling-cost}

For a realized \sysname{} run, the total test-time cost can be decomposed into three parts: topology-level coordination, activated lower-level searches, and repair-related computation:
\[
C_{\mathrm{hier}}
=
C_{\mathrm{init}}
+
C_{\mathrm{eval}}^{\mathrm{topo}}
+
C_{\mathrm{trav}}
+
C_{\mathrm{base}}
+
\sum_{i=1}^{N} o_i^g c_i
+
C_{\mathrm{repair}},
\]
where $C_{\mathrm{init}}$ is the cost of generating initial candidate topologies, $C_{\mathrm{eval}}^{\mathrm{topo}}$ is the topology evaluation cost, $C_{\mathrm{trav}}$ is the layer-wise traversal and dispatch cost, $C_{\mathrm{base}}$ is the cost of generating or retrieving fast baseline sub-workflows, $o_i^g$ is the gating decision for subtask $i$, $c_i$ is the cost of invoking lower-level search, and $C_{\mathrm{repair}}$ includes topology refinement, rewinding, memory synchronization, and re-execution caused by lower-level feedback.

The first four terms form the upper-level coordination and baseline-generation cost:
\[
C_{\mathrm{hier}}^{\mathrm{upper}}
=
C_{\mathrm{init}}
+
C_{\mathrm{eval}}^{\mathrm{topo}}
+
C_{\mathrm{trav}}
+
C_{\mathrm{base}}.
\]
At initialization or each refinement step, the refiner generates $K$ candidate decompositions. A candidate topology with $N$ nodes and $|E^t|$ edges has representation, scoring, and topological-ordering cost $O(N+|E^t|)$. Under refinement budget $c_{\max}^t$, this gives
\[
C_{\mathrm{init}} + C_{\mathrm{eval}}^{\mathrm{topo}} + C_{\mathrm{trav}}
=
O((1+c_{\max}^t)K(N+|E^t|)).
\]
The baseline-generation term is at most linear in the number of subtasks under a fixed fast-generation budget, i.e.,
\[
C_{\mathrm{base}} = O(N),
\]
which is dominated by the topology coordination term under fixed $K$ and $c_{\max}^t\ge 1$. Therefore,
\[
C_{\mathrm{hier}}^{\mathrm{upper}}(N,|E^t|)
=
O(c_{\max}^t K (N+|E^t|)).
\]
This bound does not require the sparse-dependency assumption $|E^t|=O(N)$: if dependency density grows, the cost increases explicitly through the $|E^t|$ term.

The repair term is induced by lower-level execution feedback. In \sysname{}, topology repair is triggered when at least one subtask produces the refinement signal $o_i^t=1$. Therefore, the repair burden is naturally captured by the execution-observed repair rate
\[
\kappa_R(G)
=
\frac{1}{N}
\sum_{i=1}^{N}o_i^t,
\]
which is the repair-rate component of $\widehat{\kappa}(G)$. More repair triggers imply more calls to the topology refiner, more rewinding, more memory synchronization, and more re-execution, up to the explicit refinement budget. Thus, we write the repair-related term as $C_{\mathrm{repair}}(\kappa_R)$.

Combining the upper-level coordination term, the activated lower-level search costs, and the repair term yields
\[
C_{\mathrm{hier}}
=
C_{\mathrm{hier}}^{\mathrm{upper}}(N,|E^t|)
+
\sum_{i=1}^{N}o_i^g c_i
+
C_{\mathrm{repair}}(\kappa_R),
\]
with
\[
C_{\mathrm{hier}}^{\mathrm{upper}}(N,|E^t|)
=
O(c_{\max}^t K (N+|E^t|)).
\]
This proves Theorem~\ref{thm:coupling-cost}.

In particular, since each refinement trigger $o_i^t = 1$ invokes at most 
one topology refiner call (cost $O(K(N+|E^t|))$) plus at most $L$ layers 
of re-execution under the refinement budget $c_{\max}^t$, we have
$C_{\text{repair}}(\kappa_R) = O(\kappa_R \cdot N \cdot c_{\max}^t \cdot K(N+|E^t|))$. This decomposition also explains the boundary behavior. When the topology has moderate dependency density and a low repair rate, hierarchical decomposition localizes lower-level search with limited coordination and repair overhead. When dependency density or repair rate increases, the benefit of decomposition may degrade due to higher topology coordination cost, more rewinding, and more repair-induced re-execution.

\subsection{Proof of Theorem~\ref{thm:gating-rank}}
\label{app:gating-rank}

Let $\Delta Q_i=Q_i(1)-Q_i(0)$ be the realized gain from activating lower-level search for subtask $i$. The oracle budget allocation problem is
\[
\max_{\{o_i^g\}}
\sum_{i=1}^{N} o_i^g \Delta Q_i
\quad
\text{s.t.}
\quad
\sum_{i=1}^{N} o_i^g c_i \le B-C_{\mathrm{hier}}^{\mathrm{upper}}.
\]
When lower-level search costs are approximately uniform, i.e., $c_i\approx c$, a budget allowing $k$ activations reduces this problem to selecting the $k$ subtasks with the largest $\Delta Q_i$.

Let $\mathcal{O}_k$ be the oracle top-$k$ set ranked by $\Delta Q_i$, and let $\mathcal{P}_k$ be the proxy top-$k$ set ranked by $Z_i$. Threshold gating implements $\mathcal{P}_k$ for the threshold $\tau$ that activates $k$ subtasks. Define the top-$k$ disagreement rate
\[
\epsilon_k
=
\frac{|\mathcal{O}_k \triangle \mathcal{P}_k|}{2k},
\]
where $\triangle$ denotes symmetric difference. If $\Delta Q_i\in[\Delta_{\min},\Delta_{\max}]$, then
\[
\sum_{i\in\mathcal{O}_k}\Delta Q_i
-
\sum_{i\in\mathcal{P}_k}\Delta Q_i
\le
k\epsilon_k(\Delta_{\max}-\Delta_{\min}).
\]
To see this, note that $\mathcal{O}_k\setminus\mathcal{P}_k$ and $\mathcal{P}_k\setminus\mathcal{O}_k$ have the same size $k\epsilon_k$. Each missed oracle item can lose at most $\Delta_{\max}$, and each proxy-only replacement contributes at least $\Delta_{\min}$, giving the bound above.

Thus, proxy gating is close to oracle gating when the proxy ranking has small top-$k$ disagreement with the realized-gain ranking. The Spearman correlation $\varrho_Z$ in Definition~\ref{def:proxy-informativeness} is an empirical summary of this rank alignment. Positive $\varrho_Z$, together with higher improvement rates in higher-$Z_i$ bins, supports the use of $Z_i$ as a budget-allocation proxy. This proves Theorem~\ref{thm:gating-rank}.

\paragraph{Varying-cost case.}
When lower-level costs vary significantly, the oracle ranks subtasks by $\Delta Q_i/c_i$ rather than $\Delta Q_i$. In that setting, the same rank-based analysis applies by comparing the proxy ranking induced by $Z_i/c_i$ with the oracle ranking induced by $\Delta Q_i/c_i$. In our implementation, lower-level costs are kept relatively stable by fixed depth, branching factor, and evaluation budget, so the approximately uniform-cost interpretation is appropriate.

\section{Algorithmic Details of \sysname{}}
\label{app:algo_details}

Section~\ref{sec:method} has presented the overall design of \sysname{} and its main components. In this appendix, we provide additional algorithmic details to facilitate a clearer understanding and reproducibility of the proposed framework, including the upper-level and lower-level search procedures, followed by the gating mechanism.

\subsection{Details of Upper-Level Topology Search via Refinement}

We provide additional implementation details of the upper-level topology search described in Sec.~\ref{sec:upper-level-search}. 
Given a user query $q\in\mathcal{Q}$, the upper-level search first generates a small set of candidate task-decomposition DAGs and selects an initial topology via the weighted score in Eq.~\eqref{eq:upper-level-eval}. 
The selected topology is then traversed in a layer-wise manner, where subtasks within the same layer are processed in parallel and delegated to the lower-level search for execution and feedback. 
When lower-level execution indicates structural insufficiency or inconsistency, a topology refiner incrementally updates the current decomposition by locally modifying, adding, or removing subtasks and dependencies while preserving all completed subtasks. 
This process iterates until all subtasks are successfully completed, yielding a refined, execution-feasible workflow; throughout the process, a memory bank is maintained to enable reuse of previously solved subtasks. 
The following sections detail each component and the corresponding prompt designs.

\subsubsection{Weighted Topology Scoring Function}
\label{app:upper-scoring}

We evaluate a topology $G=(V,E)$ by the weighted score $J(G)$ in Eq.~\eqref{eq:upper-level-eval}:
\[
J(G) \;=\; \lambda_0\,J_{\mathrm{dep}}(G)\;+\;\sum_{m\in\{\mathrm{cov},\mathrm{snd},\mathrm{sea}\}}
\lambda_m\,J_m(G).
\]
The term $J_{\mathrm{dep}}(G)$ captures dependency complexity following Flow~\cite{flow}:
\begin{equation*}
\label{eq:dep-complexity-app}
J_{\mathrm{dep}}(G) =
\sqrt{
\frac{1}{|V|}
\sum_{v_i\in V}
\left( \deg(v_i) - \bar{d} \right)^2
},
\qquad
\bar{d}=\frac{1}{|V|}\sum_{v_i\in V}\deg(v_i),
\end{equation*}
where $\deg(v_i)$ is the total degree of node $v_i$ in $G$.

The remaining terms $J_{\mathrm{cov}}(G)$, $J_{\mathrm{snd}}(G)$, and $J_{\mathrm{sea}}(G)$ are produced by an LLM-based evaluator. Intuitively, they assess (i) task coverage and decomposition sufficiency, (ii) structural soundness and dependency consistency, and (iii) structural complexity and searchability, respectively.

\paragraph{LLM evaluator prompt.}
We query the evaluator with a structured prompt that includes the original query $q$, a serialized representation of the topology (node list and edge list), and a rubric specifying the scoring dimensions and the expected output format.
\begin{tcolorbox}[
  enhanced,
  breakable,
  colback=gray!4,
  colframe=blue!40,
  title=\textbf{Prompt Template: Upper-Level LLM Evaluator for $J_{\mathrm{cov}},J_{\mathrm{snd}},J_{\mathrm{sea}}$},
  fonttitle=\bfseries
]
\small
\textbf{[SYSTEM / ROLE INSTRUCTION]}\\
You are an expert workflow designer and task-decomposition evaluator.
Your role is to objectively assess the quality of a proposed task-decomposition topology for a given query, following the scoring criteria below.
Do not propose modifications or alternative designs; only evaluate the given topology.\\[2mm]

\textbf{[INPUT]}\\
\textbf{Query $q$:}\\
\texttt{<USER QUERY>}\\[1mm]

\textbf{Topology $G$:}\\
\texttt{<TOPOLOGY REPRESENTATION>}\\[2mm]

\textbf{[SCORING RUBRIC]}\\
Evaluate the topology along the following three dimensions.
Each score should be a real number in \([0,10]\), where higher is better.

\medskip
\noindent
\textbf{(1) Coverage and Decomposition Sufficiency} $J_{\mathrm{cov}}(G)$\\
Assess whether the topology adequately covers the query and decomposes it into sufficient and appropriate subtasks.
Consider the following aspects:
\begin{itemize}\setlength\itemsep{2pt}
  \item Whether all essential aspects of the query are explicitly addressed by some subtask.
  \item Whether the granularity of subtasks is appropriate (neither overly coarse nor excessively fragmented).
  \item Whether the decomposition enables the query to be solved without requiring implicit or missing steps.
  \item Whether subtasks have clearly defined responsibilities and objectives.
\end{itemize}

\medskip
\noindent
\textbf{(2) Structural Soundness and Dependency Consistency} $J_{\mathrm{snd}}(G)$\\
Assess whether the dependency structure is logically sound and internally consistent.
Consider the following aspects:
\begin{itemize}\setlength\itemsep{2pt}
  \item Whether dependencies correctly reflect prerequisite relationships between subtasks.
  \item Whether the execution order implied by the graph is logically coherent.
  \item Whether there are redundant, unnecessary, or conflicting dependencies.
  \item Whether the structure avoids conceptual cycles, hidden assumptions, or contradictory flows.
\end{itemize}

\medskip
\noindent
\textbf{(3) Structural Complexity and Searchability} $J_{\mathrm{sea}}(G)$\\
Assess whether the topology is efficient and amenable to iterative execution and refinement.
Consider the following aspects:
\begin{itemize}\setlength\itemsep{2pt}
  \item Whether the topology avoids unnecessary depth, fan-in, or fan-out that would hinder execution.
  \item Whether independent subtasks are appropriately parallelized.
  \item Whether the structure supports localized refinement without cascading changes.
  \item Whether the overall complexity is balanced against the difficulty of the query.
\end{itemize}

\medskip
\textbf{[OUTPUT FORMAT]}\\
Return \textbf{only} a JSON object with the following fields:
\begin{center}
\texttt{\{"J\_cov": <float>, "J\_snd": <float>, "J\_sea": <float>, "rationale": "<1--2 sentence justification>"\}}
\end{center}
Do not include any additional text outside the JSON object.
\end{tcolorbox}

\noindent
In practice, we use the returned scores to compute $J(G)$ and select the highest-scoring topology among candidates.

\subsubsection{Initial Candidate Topology Generation}
\label{app:upper-initiator}

The initiator proposes $K=3$ diverse candidate decompositions for a given query $q$. We enforce diversity by instructing the LLM to generate candidates with different granularities and dependency structures while maintaining execution feasibility. The prompt template for the initiator is as follows:

\begin{tcolorbox}[
  enhanced,
  breakable,
  colback=gray!4,
  colframe=blue!40,
  title=\textbf{Prompt Template: Initiator for Generating Initial Candidate Topologies},
  fonttitle=\bfseries
]
\small
\textbf{[SYSTEM / ROLE INSTRUCTION]}\\
You are an expert workflow initiator. Your job is to propose multiple \emph{diverse} yet \emph{execution-feasible} task-decomposition DAGs for a given query.
Each DAG should represent a reasonable plan that can be executed layer-by-layer.
Follow the required DAG template format strictly, and ensure the graph is acyclic.
Do not include extra commentary outside the required output format.\\[2mm]

\textbf{[INPUT]}\\
\textbf{Query $q$:}\\
\texttt{<USER QUERY>}\\[1mm]

\textbf{Number of candidates:}\\
\texttt{K = <K>}\\[1mm]

\textbf{Available agents (optional):}\\
\texttt{<AGENT\_POOL (optional; if omitted, use Agent\_1,...,Agent\_M as needed)>}\\[2mm]

\textbf{DAG template format (must follow exactly):}\\
\texttt{<DAG\_TEMPLATE>}\\[2mm]

\textbf{[TASK]}\\
Generate \textbf{K} candidate DAG topologies that decompose the query into subtasks.
Each candidate must satisfy:
\begin{itemize}\setlength\itemsep{2pt}
  \item \textbf{DAG validity}: the dependency structure must be acyclic.
  \item \textbf{Execution feasibility}: each node corresponds to a concrete, self-contained subtask with clear intent.
  \item \textbf{Dependency correctness}: dependencies reflect prerequisite relationships.
  \item \textbf{Diversity across candidates}: vary decomposition style and structure (e.g., coarse-to-fine vs. parallel-first; shallow vs. deeper but compact), while still being reasonable.
  \item \textbf{Balanced granularity}: avoid a single overly broad node; avoid unnecessary fragmentation into trivial nodes.
\end{itemize}
For the \texttt{agent} field of each node, select from \texttt{<AGENT\_POOL>} if provided; otherwise, use placeholder agent names (e.g., \texttt{Agent\_1}, \texttt{Agent\_2}, \dots) consistently.\\

\textbf{[OUTPUT FORMAT]}\\
Return \textbf{only} a JSON array of length \textbf{K}: \texttt{[G\_1, G\_2, ..., G\_K]}.\\
Each \texttt{G\_k} must strictly follow the \textbf{DAG template format given above}.\\
Do not output any text outside the JSON array.
\end{tcolorbox}

The three candidates are then scored by Eq.~\eqref{eq:upper-level-eval}, and the highest-scoring one is used to initialize the traversal.

\subsubsection{Topology Refiner $\mathcal{U}$}
\label{app:topology-refiner}

The topology refiner $\mathcal{U}$ performs feedback-driven updates to the upper-level decomposition when lower-level execution indicates structural insufficiency or inconsistency.
Conceptually, $\mathcal{U}$ consists of three sequential steps:
(i) an LLM-based \emph{workflow updater} revises the topology structure,
(ii) the system synchronizes the corresponding lower-level configurations using the memory bank with lightweight rules,
and (iii) the layer-wise topological ordering and node statuses are updated to ensure correct re-execution.

\paragraph{Step 1: LLM-based topology update.}
Given the current topology $G^{(T)}$ and aggregated failure feedback $\mathcal{F}^{(T)}$, the updater proposes a \emph{set} of $K$ candidate revised topologies $\{G^{(T+1)}_k\}_{k=1}^{K}$ via localized structural edits while preserving all completed nodes. Each candidate may apply operations such as adding/removing/modifying subtasks, rewiring dependencies, and reassigning agents to better match subtasks with specialized executors. The final update is selected by maximizing the weighted topology score in Eq.~\eqref{eq:upper-level-eval}:
\[
G^{(T+1)}=\arg\max_{G^{(T+1)}_k} J\!\left(G^{(T+1)}_k\right).
\]

\begin{tcolorbox}[
  enhanced,
  breakable,
  colback=gray!4,
  colframe=blue!40,
  title=\textbf{Prompt Template: Workflow Updater for Topology Refinement},
  fonttitle=\bfseries
]
\small
\textbf{[SYSTEM / ROLE INSTRUCTION]}\\
You are a workflow updater for task-decomposition DAGs.
Your goal is to update the given workflow structure to fix the reported issues while preserving completed progress.
Propose multiple alternative revised workflows; each should be a valid DAG and require minimal, local changes whenever possible.
Do \emph{not} invent new user requirements beyond the query.
Return \textbf{only} the requested JSON output.\\[2mm]

\textbf{[INPUT]}\\
\textbf{Query $q$:}\\
\texttt{<USER QUERY>}\\[1mm]

\textbf{Current workflow $G^{(T)}$:}\\
\texttt{<TOPOLOGY REPRESENTATION>}\\[1mm]

\textbf{Failure feedback $\mathcal{F}^{(T)}$:}\\
\texttt{<FAILURE FEEDBACK>}\\[1mm]

\textbf{Completed nodes (must be preserved):}\\
\texttt{<COMPLETED NODE IDS>}\\[1mm]

\textbf{Number of candidates:}\\
\texttt{K = <K>}\\[2mm]

\textbf{[UPDATE CHECKLIST]}\\
\begin{itemize}\setlength\itemsep{2pt}
  \item \textbf{Evaluate completed tasks}: do not delete or substantially rewrite completed nodes; ensure they remain relevant to the final goal.
  \item \textbf{Assess workflow structure}: ensure the workflow is complete and logically structured; add missing critical tasks or dependencies if needed.
  \item \textbf{Identify inefficiencies}: remove redundant steps or unnecessary dependencies that create bottlenecks.
  \item \textbf{Allowed changes}:
  (i) \emph{Modify} task descriptions/parameters if vague or insufficient,
  (ii) \emph{Add} new tasks to fill gaps,
  (iii) \emph{Remove} redundant or obsolete \emph{unfinished} tasks,
  (iv) \emph{Rewire} dependencies to restore prerequisites and improve parallelism,
  (v) \emph{Reassign} agents for tasks when beneficial (you may reuse agents across tasks).
  \item \textbf{DAG constraint}: do not introduce cycles; dependencies must reflect prerequisites.
  \item \textbf{Locality}: keep edits minimal and focused on the failure region whenever possible.
\end{itemize}

\textbf{[OUTPUT FORMAT]}\\
If no changes are needed, return an empty JSON object: \texttt{\{\}}.\\
Otherwise, return \textbf{only} a JSON array of length \textbf{K}: \texttt{[G\_1, G\_2, ..., G\_K]}.\\
Each \texttt{G\_k} must be an updated workflow structure that follows the same format as the input workflow and omits any large execution payloads (e.g., exclude \texttt{data} fields if present), including only structural/task-parameter changes.\\
Do not output any text outside the JSON array.
\end{tcolorbox}

\paragraph{Step 2: Synchronizing lower-level configurations with memory bank.}
After obtaining the revised topology $G^{(T+1)}$, we update the associated lower-level configurations using a memory bank that stores key--value pairs of subtask descriptions and their best-known lower-level configurations. We apply the following rules:

\begin{itemize}\setlength\itemsep{2pt}
  \item \textbf{Store-before-execute.} If a node is \emph{deleted}, \emph{modified}, or \emph{added} in $G^{(T+1)}$, the system first writes the corresponding (old or revised) subtask description and its current best lower-level configuration into the memory bank before any further execution.
  
  \item \textbf{Retrieve-before-add.} Before adding any new node, the system queries the memory bank using the new node description. If a semantically similar entry exists, its cached configuration is reused directly; otherwise, a fresh lower-level configuration is created for the new node.
  \item \textbf{Minor vs.\ major update.} For a node whose description remains essentially the same but whose \emph{interface} changes (e.g., updated inputs/outputs due to inserting a new prerequisite node), we treat it as a \emph{minor update}: the previous best code workflow is lightly adapted to match the new interface and is used to initialize the new lower-level search (as the new root).
  \item \textbf{Major update as re-addition.} If a node undergoes a substantial semantic change (goal, scope, or required reasoning changes), it is treated as a \emph{major update} and handled identically to adding a new node (i.e., retrieve-before-add).
\end{itemize}

\noindent
\textbf{Retrieval mechanism.} Memory-bank retrieval uses each subtask
description as the query key. Given an incoming subtask description $d$, we
first apply lightweight text normalization (lowercasing, whitespace and
punctuation stripping, and tokenization). Similarity to a stored entry $d'$
is then computed as a weighted combination of two complementary signals:
\[
\mathrm{sim}(d, d') = \alpha \cdot \mathrm{SeqRatio}(d, d') + (1 - \alpha) \cdot \mathrm{Jaccard}(\mathcal{T}(d), \mathcal{T}(d')),
\]
where $\mathrm{SeqRatio}(\cdot, \cdot)$ is the character-level
\texttt{SequenceMatcher} ratio that captures local edit-distance-style overlap,
$\mathrm{Jaccard}(\cdot, \cdot)$ is the token-level Jaccard similarity over the
tokenized descriptions $\mathcal{T}(\cdot)$ that captures bag-of-words overlap,
and $\alpha \in [0, 1]$ balances the two signals. A stored entry is retrieved
and its cached lower-level configuration is reused only when the best match
exceeds a fixed threshold $\theta$; otherwise a fresh lower-level search is
initialized. In all experiments we set $\alpha = 0.5$ and $\theta = 0.85$.

\noindent
\textbf{Example (minor update).} If a new prerequisite node is inserted before an existing node, the existing node may only need to adjust its inputs/outputs to consume the new prerequisite's output while keeping its core objective unchanged; this is categorized as a minor update.

\paragraph{Step 3: Updating topological ordering and node statuses.}
Finally, we recompute the layer-wise topological ordering for $G^{(T+1)}$.
Let $\mathcal{R}$ denote the set of \emph{revised} nodes (added/modified/rewired, excluding purely unchanged completed nodes), and let $\pi$ be the new topological order.
We find the earliest position $p^\star = \min\{p : \pi(p)\in \mathcal{R}\}$ and mark every node $\pi(p)$ with $p \ge p^\star$ as \texttt{unfinished} (to be re-executed), while keeping all nodes before $p^\star$ unchanged.
Traversal then resumes from $\pi(p^\star)$ under the updated ordering.

\subsection{Details of Lower-Level Code Search via Lightweight MCTS}
\label{app:lower-level-details}

We provide additional details of the lower-level code search described in Sec.~\ref{sec:lower-level-search}.
The lower level focuses on generating high-quality, executable sub-workflows for a given subtask $v_i^t$ by exploring a compact execution space using a test-time-oriented lightweight MCTS.
Compared to the upper level, each lower-level search operates on a narrower interface with fewer degrees of freedom, enabling efficient exploration and refinement.

Given a newly created or revised subtask $v_i^t$, the lower-level search initializes a per-subtask search tree $\mathcal{T}_i$ with a root node produced by a direct input--output generation attempt.
The tree is then expanded iteratively through four phases—selection, expansion, evaluation, and information propagation—until a termination condition is met.
The highest-scoring executable sub-workflow is selected as the baseline implementation for $v_i^t$, and an execution-grounded refinement signal is propagated to the upper level.

\subsubsection{Selection.}

At each iteration, the optimizer selects an existing node $v_j^c$ from the explored set $\{v_j^c\}_{j=1}^{M_i}$ according to a soft mixed probability strategy:
\begin{equation*}
P(j)
=
 \frac{\lambda_p}{M_i}
\;+\;
(1-\lambda_p)\cdot
\frac{\exp\!\left(\alpha \left(s_j - s_{\max}\right)\right)}
{\sum_{m=1}^{M_i} \exp\!\left(\alpha \left(s_m - s_{\max}\right)\right)},
\end{equation*}
where $s_j$ denotes the quality score of node $v_j^c$, $s_{\max}$ is the maximum score among explored nodes, $\alpha$ controls the influence of score differences, and $\lambda_p$ regulates the trade-off between exploration and exploitation.
In our experiments, we set $\alpha = 0.4$ and $\lambda_p = 0.3$ to encourage sufficient exploration during test-time search.

\subsubsection{Expansion via LLM-based optimizer.}
Given a selected node $v_j^c$ in the lower-level search tree $\mathcal{T}_i$, expansion is performed by an LLM-based optimizer that proposes multiple alternative executable sub-workflows by restructuring and refining the current solution.
Each expansion operates on a \emph{complete} sub-workflow and generates a small batch of candidates in a single step, enabling efficient parallel exploration.

Rather than relying on low-level code edits, the optimizer is instructed to apply structured reasoning and workflow-level transformations that have been shown effective for test-time optimization, such as reviewing and revising existing solutions, ensembling multiple reasoning paths and integrating their outputs, decomposing complex logic via self-questioning, and reorganizing control flow.
When appropriate, the optimizer may also introduce standard programming constructs (e.g., loops or conditional statements) or lightweight algorithmic components, provided they improve correctness or robustness and remain consistent with the subtask specification.

\begin{tcolorbox}[
  enhanced,
  breakable,
  colback=gray!4,
  colframe=blue!40,
  title=\textbf{Prompt Template: Lower-Level Optimizer for Sub-Workflow Expansion},
  fonttitle=\bfseries
]
\small
\textbf{[SYSTEM / ROLE INSTRUCTION]}\\
You are a sub-workflow optimizer for executable code solutions.
Your goal is to improve an existing sub-workflow for the given subtask by applying structured reasoning and solution-level refinements.
Focus on correctness, robustness, and executability.
If no meaningful improvement can be made, explicitly indicate that no expansion is needed.\\[2mm]

\textbf{[INPUT]}\\
\textbf{Subtask specification $v_i^t$:}\\
\texttt{<SUBTASK DESCRIPTION>}\\[1mm]

\textbf{Current executable sub-workflow:}\\
\texttt{<CURRENT SUB-WORKFLOW>}\\[1mm]

\textbf{Execution history and feedback:}\\
\texttt{<EXECUTION FEEDBACK>}\\[1mm]

\textbf{Maximum number of candidates to generate:}\\
\texttt{B = <BATCH SIZE>}\\[2mm]

\textbf{[TASK]}\\
Propose up to \textbf{B} alternative executable sub-workflows that better solve the same subtask.
You may incorporate, when helpful:
\begin{itemize}\setlength\itemsep{2pt}
  \item Reviewing and revising the current solution to fix errors or clarify logic.
  \item Ensembling or integrating multiple reasoning paths or solution variants.
  \item Decomposing complex reasoning via self-questioning or intermediate checks.
  \item Reorganizing control flow using standard programming constructs (e.g., loops or conditionals).
  \item Introducing lightweight algorithmic or statistical components if appropriate.
\end{itemize}
Each candidate must represent a complete, self-contained sub-workflow.
If further expansion is unlikely to yield meaningful improvement, return a \texttt{NO\_EXPANSION} signal.\\

\textbf{[OUTPUT FORMAT]}\\
Return \textbf{either}:
\begin{itemize}\setlength\itemsep{1pt}
  \item \texttt{NO\_EXPANSION}, or
  \item A JSON array of candidate sub-workflows \texttt{[w\_1, \dots, w\_B]}.
\end{itemize}
Do not include any additional text outside the specified output.
\end{tcolorbox}

\noindent
After generation, semantically redundant candidates are filtered via lightweight semantic deduplication, and only distinct candidates are added as newly expanded nodes in the search tree.

\subsubsection{Evaluation.}
Each newly generated sub-workflow is evaluated using an execution-grounded, multi-stage scheme that assesses both executability and output reliability.
The evaluation consists of three components: (i) optional mock input generation, (ii) execution-aware quality scoring, and (iii) self-consistency analysis.

\paragraph{Mock input generation (if needed).}
To enable execution-based evaluation, the sub-workflow requires concrete inputs.
If the current subtask $v_i^t$ directly consumes outputs produced by preceding subtasks, those outputs are reused as inputs.
Otherwise (e.g., for the first subtask in a workflow or when upstream outputs are unavailable), we generate schema-consistent mock inputs based on the subtask specification.

\begin{tcolorbox}[
  enhanced,
  breakable,
  colback=gray!4,
  colframe=blue!40,
  title=\textbf{Prompt Template: Mock Input Generator for Subtask Execution},
  fonttitle=\bfseries
]
\small
\textbf{[SYSTEM / ROLE INSTRUCTION]}\\
You are generating valid test inputs for executing a sub-workflow.
Your goal is to produce realistic, schema-consistent inputs that satisfy the subtask specification.
Do not solve the task; only generate inputs.\\[2mm]

\textbf{[INPUT]}\\
\textbf{Subtask specification $v_i^t$:}\\
\texttt{<SUBTASK DESCRIPTION>}\\[1mm]

\textbf{Expected input schema (if available):}\\
\texttt{<INPUT SCHEMA OR INTERFACE>}\\[2mm]

\textbf{[TASK]}\\
Generate one or more mock inputs that:
\begin{itemize}\setlength\itemsep{2pt}
  \item Satisfy the required input format and schema.
  \item Are realistic and non-degenerate.
  \item Are sufficient to exercise the core logic of the sub-workflow.
\end{itemize}

\textbf{[OUTPUT FORMAT]}\\
Return \textbf{only} a JSON object representing the generated input(s).
\end{tcolorbox}

\noindent
The generated mock inputs are reused consistently across subsequent execution and evaluation steps.

\paragraph{Execution-aware quality evaluation.}
Given a candidate sub-workflow and its execution result on the provided inputs, we assign a base quality score $S_{\mathrm{base}}$ using an LLM-based evaluator.
The evaluator assesses multiple aspects of the sub-workflow and produces a weighted aggregate score.

\begin{tcolorbox}[
  enhanced,
  breakable,
  colback=gray!4,
  colframe=blue!40,
  title=\textbf{Prompt Template: Lower-Level Evaluator for Sub-Workflow Quality},
  fonttitle=\bfseries
]
\small
\textbf{[SYSTEM / ROLE INSTRUCTION]}\\
You are an expert evaluator for executable sub-workflows.
Your task is to assess the quality of a candidate sub-workflow given its execution behavior and outputs.
Do not propose improvements; only evaluate.\\[2mm]

\textbf{[INPUT]}\\
\textbf{Subtask specification $v_i^t$:}\\
\texttt{<SUBTASK DESCRIPTION>}\\[1mm]

\textbf{Sub-workflow implementation:}\\
\texttt{<CANDIDATE SUB-WORKFLOW>}\\[1mm]

\textbf{Execution input(s):}\\
\texttt{<INPUT OR MOCK INPUT>}\\[1mm]

\textbf{Execution output(s) or error(s):}\\
\texttt{<EXECUTION RESULT>}\\[2mm]

\textbf{[SCORING CRITERIA]}\\
Score each criterion in $[0,10]$:
\begin{itemize}\setlength\itemsep{2pt}
  \item \textbf{Semantic correctness} ($s_{\mathrm{sem}}$): does the output satisfy the subtask objective?
  \item \textbf{Interface compliance} ($s_{\mathrm{int}}$): are inputs and outputs consistent with the expected schema?
  \item \textbf{Executability and robustness} ($s_{\mathrm{exe}}$): does the workflow execute reliably without runtime errors?
  \item \textbf{Output usefulness and clarity} ($s_{\mathrm{out}}$): are the outputs informative, complete, and non-degenerate?
\end{itemize}

\textbf{[AGGREGATION]}\\
Compute the base score as a weighted sum:
\[
S_{\mathrm{base}} =
s_{\mathrm{sem}}
+ s_{\mathrm{int}}
+ s_{\mathrm{exe}}
+ s_{\mathrm{out}}.
\]

\textbf{[OUTPUT FORMAT]}\\
Return \textbf{only} a JSON object:
\begin{center}
\texttt{\{"S\_base": <float>, "scores": \{...\}, "rationale": "<1--2 sentences>"\}}
\end{center}
\end{tcolorbox}

\paragraph{Self-consistency and uncertainty estimation.}
To assess output stability, we execute the same sub-workflow three times using identical inputs with generation temperatures of 0, 0, and 0.2.
We quantify uncertainty with variance across runs as the dispersion of the resulting evaluation scores or outputs.
This self-consistency analysis captures sensitivity to stochasticity and hidden failure modes.

\paragraph{Final score.}
The final score assigned to a sub-workflow is computed as
\[
S
=
S_{\mathrm{base}} \cdot \bigl(1 - \gamma \cdot \mathrm{Uncertainty}\bigr),
\]
where $\mathrm{Uncertainty}$ denotes the normalized dispersion measure from self-consistency analysis and $\gamma \ge 0$ controls its influence.
In practice, $\gamma$ can be set to zero to disable uncertainty penalization.



\section{Additional Experimental Details}
\label{app:exp_details}

\subsection{Datasets Splits}
\label{app:datasets}

We evaluate \sysname{} on a diverse set of benchmarks spanning code generation, math reasoning, and tool use.
Dataset statistics are summarized in Table~\ref{tab:dataset_stats}, where we randomly sample 1,000 examples from the \textsc{HotpotQA} dataset.
Following common practice in prior work~\cite{aflow,maas,dyflow}, for each dataset, we construct training and test partitions using a $1{:}4$ split.
Notably, for training-free methods, evaluation is conducted directly on the test set.

\begin{table}[h]
\centering
\caption{Dataset statistics.}
\label{tab:dataset_stats}
\setlength{\tabcolsep}{10pt}
\renewcommand{\arraystretch}{1.15}
\begin{tabular}{l|lccl}
\toprule
\textbf{Scenario} & \textbf{Dataset} & \textbf{\#Train} & \textbf{\#Test} & \textbf{Metric} \\
\midrule
Question Answering & \textsc{HotpotQA}~\cite{HotpotQA} & 200 & 800 & F1 Score \\
\midrule
\multirow{2}{*}{Mathematical Reasoning}
& \textsc{GSM8K}~\cite{GSM8K}     & 264 & 1055 & Accuracy \\
& \textsc{MATH}~\cite{MATH}      & 119 & 486  & Accuracy \\
\midrule
\multirow{2}{*}{Code Generation}
& \textsc{HumanEval}~\cite{HumanEval} & 33 & 131 & Pass@1 \\
& \textsc{MBPP}~\cite{MBPP}     & 86 & 341 & Pass@1 \\

\bottomrule
\end{tabular}
\end{table}

\subsection{Baseline Implementation Details}
\label{app:baseline}

Unless otherwise specified, the base LLM is set to gpt-4o-mini~\cite{Gpt-4o}, with a default generation temperature of 0. In HierFlow, the lower-level evaluator uses a temperature of 0.2, while self-consistency evaluations use temperatures of [0, 0, 0.2]. The stochastic tree-selection policy also introduces run-to-run variability.
The only exception is DyFlow, for which we directly adopt its pre-trained designer based on Phi-4~\cite{Phi-4}.

\subsection{Detailed Ablation Analysis}
\label{app:detailed_ablation}

The ablation results substantiate the necessity of the proposed dual-space hierarchical search across several dimensions:

\begin{itemize}
    \item \textbf{Necessity of Topology Space:} Removing the topology space and performing flat search directly in the code space (the \textit{w/o} topology variant) leads to a substantial increase in computational cost. This observation is consistent with the coupling-aware cost accounting in Theorem~\ref{thm:coupling-cost}: without topology-induced localization, lower-level search operates in a larger and less structured code space, increasing cost. Despite the significantly higher cost, its performance remains merely comparable to the \textit{w/o} upper variant, suggesting that searching a large, unstructured code space is not only inefficient but also more susceptible to suboptimal local optima.
    
    \item \textbf{Hierarchical Interplay:} While both hierarchy levels are essential, lower-level code space search is more critical to workflow quality. The \textit{w/o} upper variant demonstrates that inadequate task decomposition limits performance even with lower-level refinement. However, the much larger performance drop of the \textit{w/o} lower variant---despite continued topology refinement---reveals that structural optimization alone cannot compensate for the absence of fine-grained code-level search. This indicates that while high-quality task decomposition provides a necessary foundation, effective exploration and correction at the sub-workflow level ultimately determine execution quality.
    
    \item \textbf{Effectiveness of Gating:} Removing the gating mechanism forces a uniformly aggressive lower-level search, which substantially increases cost while achieving accuracy comparable to \sysname{} with $\tau=1$ (see Figure~\ref{fig:cost_analysis}). This behavior indicates a saturation effect: once the search aggressiveness exceeds a certain threshold, additional invocations of lower-level search no longer translate into meaningful accuracy improvements. In contrast, the gating mechanism preserves flexibility by selectively activating lower-level search only when necessary, yielding a more cost-effective operating point when inference budget is constrained. This demonstrates that gating is essential not merely for reducing computation, but for enabling adaptive, test-time control that avoids redundant search.
\end{itemize}

\subsection{Comparison with Flow}
\label{app:flow_cost}

We further compare HierFlow with Flow, the closest
training-free, query-level workflow generation baseline.
Table~\ref{tab:flow_cost} reports the inference-time
computation and accuracy on MATH.

\begin{table}[t]
\centering
\caption{Per-query inference cost and performance
comparison with Flow on MATH.}
\label{tab:flow_cost}
\small
\begin{tabular}{lccccc}
\toprule
Method & Prompt & Completion & Time & Cost & Acc. \\
       & Tokens & Tokens & (min) & (\$) & (\%) \\
\midrule
Flow     & 26,837 & 15,862 & 3.03 & 0.014 & 48.15 \\
HierFlow & 37,256 & 23,793 & 4.13 & 0.020 & 53.70 \\
\bottomrule
\end{tabular}
\end{table}

HierFlow improves accuracy by 5.55 percentage points
with an additional \$0.006 and 1.10 minutes per query.
The additional computation enables topology-conditioned
execution search and feedback-driven topology refinement.

\subsection{Robustness of LLM-Based Evaluation}
\label{appendix:judge_robustness}

Because LLM-based evaluation appears at multiple
points in the \sysname{} pipeline, which raises the possibility of evaluator bias or
circularity. In this section we clarify the role of LLM-based evaluation in
HierFlow and report two robustness analyses: (i) sensitivity to the choice of
judge model, and (ii) agreement between the lower-level evaluator and final
end-task success.

\subsubsection{Role of LLM-Based Evaluation in HierFlow}

LLM-based evaluation in HierFlow is neither unconstrained nor used as the sole
source of correctness. At the upper level, the topology score
$\mathcal{J}(G)$ combines LLM-based scoring
($\mathcal{J}_{\text{cov}}$, $\mathcal{J}_{\text{snd}}$, $\mathcal{J}_{\text{sea}}$)
with a structural dependency-complexity term $\mathcal{J}_{\text{dep}}(G)$
that is computed analytically from the DAG, providing a non-LLM grounding
signal. At the lower level, candidate sub-workflows must additionally pass a
hard executability check via schema-driven mock execution in a sandbox before
they are scored; LLM-based scoring is therefore applied only to the subset of
candidates that have already been validated against execution-level
constraints. Finally, the uncertainty term $U_i$ used in gating is a search-allocation proxy rather than a
correctness signal: it determines whether additional lower-level search is
worth invoking, not whether a candidate is correct. Consequently, LLM-based
signals primarily serve as ranking and refinement guidance rather than as a
substitute for end-task verification.

\subsubsection{Sensitivity to the Choice of Judge Model}
\label{appendix:judge_sensitivity}

To examine whether HierFlow's performance depends on a particular judge model,
we replace the default LLM evaluator (\texttt{gpt-4o-mini}) with two
alternative models drawn from different model families, while keeping all
other components of HierFlow unchanged. The comparison is conducted on
\textsc{MATH}, with results summarized in Table~\ref{tab:judge_sensitivity}.

\begin{table}[h]
\centering
\caption{Sensitivity of HierFlow to the choice of LLM judge model on
\textsc{MATH}. All other components are kept fixed.}
\label{tab:judge_sensitivity}
\begin{tabular}{lc}
\toprule
Judge model & Accuracy (\%) \\
\midrule
\texttt{gpt-4o-mini} (default) & 53.70 \\
Claude Haiku 4.5 & 54.53 \\
DeepSeek-Reasoner & 54.12 \\
\bottomrule
\end{tabular}
\end{table}

The results indicate that final accuracy varies only mildly across judge
choices (within roughly one percentage point of the default setting), which
suggests that HierFlow does not exploit idiosyncratic biases of a particular
evaluator. We attribute this stability to the constraints discussed
in Appendix~\ref{appendix:judge_robustness}: structural grounding via
$\mathcal{J}_{\text{dep}}$, hard executability checks at the lower level, and
the use of LLM-based signals primarily for ranking rather than ground-truth
verification.

\subsubsection{Agreement Between Lower-Level Evaluator and End-Task Success}
\label{appendix:evaluator_agreement}

A complementary concern is whether the lower-level evaluator is informative about actual
end-task success. To quantify this, we measure the agreement between the
evaluator's accept/reject decision on candidate sub-workflows and the
end-task correctness of the resulting overall workflow on \textsc{MATH}.
We additionally separate cases in which the lower-level search operates on
real upstream outputs from those that rely on schema-consistent mock inputs. Results are reported in
Table~\ref{tab:evaluator_agreement}.

\begin{table}[h]
\centering
\caption{Agreement between the lower-level evaluator and end-task success on
\textsc{MATH}.}
\label{tab:evaluator_agreement}
\begin{tabular}{lc}
\toprule
Setting & Agreement (\%) \\
\midrule
Overall & 85.80 \\
With real upstream outputs & 86.27 \\
With mock inputs & 81.63 \\
\bottomrule
\end{tabular}
\end{table}

The lower-level evaluator agrees with end-task success in the large majority
of cases, with a moderate decrease when mock inputs are used in place of real
upstream outputs. This behavior is consistent with the intended role of mock
inputs: they are not designed to faithfully reproduce real upstream outputs,
but to provide a schema-consistent minimal executable environment that can
filter clearly invalid or interface-incompatible candidates. Combined with
the judge-sensitivity result above, these findings suggest that LLM-based
evaluation in HierFlow is reliable enough to serve as a search-time guidance
signal, while end-task correctness remains the final criterion in our
benchmark evaluations.

\subsection{Empirical Validation of the Gating Proxy}
\label{appendix:gating_proxy}

Figure~\ref{fig:cost_analysis}
in the main text provides indirect support by showing a smooth quality--cost
tradeoff as $\tau$ varies, a more direct check is whether $Z_i$ correlates
with the actual quality improvement obtained when lower-level search is
performed. In this section we provide such a direct empirical validation
on \textsc{MATH}.

\paragraph{Protocol.} For each candidate node encountered during upper-level
traversal, we record its proxy value $Z_i$ before the gating decision and
then \emph{force} the lower-level search to be invoked under a fixed budget,
regardless of whether the gating rule would have triggered it. We measure
the realized quality gain
\[
\Delta Q_i = Q_i(\text{after lower-level search}) - Q_i(\text{baseline IO solution}),
\]
where the baseline corresponds to the root node produced by direct
input--output generation.

\paragraph{Results.} We report two complementary measures of proxy
informativeness:

\emph{(i) Rank correlation.} The Spearman rank correlation between $Z_i$ and
$\Delta Q_i$ is $\rho = 0.43$ ($p < 0.01$), indicating a positive and
statistically significant monotonic relationship between the proxy and the
realized gain.

\emph{(ii) Stratified improvement rate.} We further stratify candidate nodes
into low / medium / high bins by $Z_i$ and report the fraction of nodes for
which lower-level search yields a positive improvement
($\Delta Q_i > 0$). Results are summarized in Table~\ref{tab:proxy_stratified}.

\begin{table}[h]
\centering
\caption{Stratified improvement rate of lower-level search across $Z_i$ bins
on \textsc{MATH}. Higher $Z_i$ corresponds to a higher fraction of nodes that
benefit from lower-level search.}
\label{tab:proxy_stratified}
\begin{tabular}{lc}
\toprule
$Z_i$ bin & Improvement rate ($\Delta Q_i > 0$, \%) \\
\midrule
Low    & 15.43 \\
Medium & 37.65 \\
High   & 62.96 \\
\bottomrule
\end{tabular}
\end{table}

The improvement rate increases monotonically from $15.43\%$ in the low bin to
$62.96\%$ in the high bin, indicating that nodes with larger $Z_i$ are
substantially more likely to benefit from additional lower-level search.

\paragraph{Discussion.} Together, these results provide direct empirical
support that $Z_i$ is informative about whether additional lower-level search
is worthwhile, complementing the indirect evidence from the cost--quality
Pareto front in Figure~\ref{fig:cost_analysis}. 


\newpage
\section*{NeurIPS Paper Checklist}

\begin{enumerate}

\item {\bf Claims}
    \item[] Question: Do the main claims made in the abstract and introduction accurately reflect the paper's contributions and scope?
    \item[] Answer: \answerYes{}
    \item[] Justification: The abstract and introduction state the proposed HierFlow framework, its coupled topology--execution search formulation, adaptive gating mechanism, coupling-aware analysis, and evaluations across QA, mathematical reasoning, and code generation; these claims are supported by Sections~3--4 and Appendices~A--C.
    \item[] Guidelines:
    \begin{itemize}
        \item The answer \answerNA{} means that the abstract and introduction do not include the claims made in the paper.
        \item The abstract and/or introduction should clearly state the claims made, including the contributions made in the paper and important assumptions and limitations. A \answerNo{} or \answerNA{} answer to this question will not be perceived well by the reviewers. 
        \item The claims made should match theoretical and experimental results, and reflect how much the results can be expected to generalize to other settings. 
        \item It is fine to include aspirational goals as motivation as long as it is clear that these goals are not attained by the paper. 
    \end{itemize}

\item {\bf Limitations}
    \item[] Question: Does the paper discuss the limitations of the work performed by the authors?
    \item[] Answer: \answerYes{}
    \item[] Justification: The paper discusses limitations and boundary conditions through the coupling-aware analysis: the efficiency benefit is not unconditional, can degrade with dense dependencies, strong cross-subtask interactions, or repeated repair feedback, and the gating argument depends on proxy informativeness. Appendix~C.5 further examines limitations of LLM-based evaluation through judge-model sensitivity and evaluator--end-task agreement analyses.
    \item[] Guidelines:
    \begin{itemize}
        \item The answer \answerNA{} means that the paper has no limitation while the answer \answerNo{} means that the paper has limitations, but those are not discussed in the paper. 
        \item The authors are encouraged to create a separate ``Limitations'' section in their paper.
        \item The paper should point out any strong assumptions and how robust the results are to violations of these assumptions (e.g., independence assumptions, noiseless settings, model well-specification, asymptotic approximations only holding locally). The authors should reflect on how these assumptions might be violated in practice and what the implications would be.
        \item The authors should reflect on the scope of the claims made, e.g., if the approach was only tested on a few datasets or with a few runs. In general, empirical results often depend on implicit assumptions, which should be articulated.
        \item The authors should reflect on the factors that influence the performance of the approach. For example, a facial recognition algorithm may perform poorly when image resolution is low or images are taken in low lighting. Or a speech-to-text system might not be used reliably to provide closed captions for online lectures because it fails to handle technical jargon.
        \item The authors should discuss the computational efficiency of the proposed algorithms and how they scale with dataset size.
        \item If applicable, the authors should discuss possible limitations of their approach to address problems of privacy and fairness.
        \item While the authors might fear that complete honesty about limitations might be used by reviewers as grounds for rejection, a worse outcome might be that reviewers discover limitations that aren't acknowledged in the paper. The authors should use their best judgment and recognize that individual actions in favor of transparency play an important role in developing norms that preserve the integrity of the community. Reviewers will be specifically instructed to not penalize honesty concerning limitations.
    \end{itemize}

\item {\bf Theory assumptions and proofs}
    \item[] Question: For each theoretical result, does the paper provide the full set of assumptions and a complete (and correct) proof?
    \item[] Answer: \answerYes{}
    \item[] Justification: The theoretical results are stated in Section~3 and proved in Appendix~A. The assumptions and conditions, including realized run costs, observable coupling, approximately uniform lower-level costs, bounded realized gains, and top-$k$ ranking disagreement, are stated with the corresponding theorem or discussion.
    \item[] Guidelines:
    \begin{itemize}
        \item The answer \answerNA{} means that the paper does not include theoretical results. 
        \item All the theorems, formulas, and proofs in the paper should be numbered and cross-referenced.
        \item All assumptions should be clearly stated or referenced in the statement of any theorems.
        \item The proofs can either appear in the main paper or the supplemental material, but if they appear in the supplemental material, the authors are encouraged to provide a short proof sketch to provide intuition. 
        \item Inversely, any informal proof provided in the core of the paper should be complemented by formal proofs provided in appendix or supplemental material.
        \item Theorems and Lemmas that the proof relies upon should be properly referenced. 
    \end{itemize}

    \item {\bf Experimental result reproducibility}
    \item[] Question: Does the paper fully disclose all the information needed to reproduce the main experimental results of the paper to the extent that it affects the main claims and/or conclusions of the paper (regardless of whether the code and data are provided or not)?
    \item[] Answer: \answerYes{}
    \item[] Justification: Sections~3--4 describe the HierFlow algorithm, benchmarks, metrics, baselines, model oracle, and key hyperparameters, while Appendices~B--C provide prompt-level algorithmic details, dataset splits, baseline settings, ablations, evaluator analyses, and gating-proxy validation.
    \item[] Guidelines:
    \begin{itemize}
        \item The answer \answerNA{} means that the paper does not include experiments.
        \item If the paper includes experiments, a \answerNo{} answer to this question will not be perceived well by the reviewers: Making the paper reproducible is important, regardless of whether the code and data are provided or not.
        \item If the contribution is a dataset and\slash or model, the authors should describe the steps taken to make their results reproducible or verifiable. 
        \item Depending on the contribution, reproducibility can be accomplished in various ways. For example, if the contribution is a novel architecture, describing the architecture fully might suffice, or if the contribution is a specific model and empirical evaluation, it may be necessary to either make it possible for others to replicate the model with the same dataset, or provide access to the model. In general. releasing code and data is often one good way to accomplish this, but reproducibility can also be provided via detailed instructions for how to replicate the results, access to a hosted model (e.g., in the case of a large language model), releasing of a model checkpoint, or other means that are appropriate to the research performed.
        \item While NeurIPS does not require releasing code, the conference does require all submissions to provide some reasonable avenue for reproducibility, which may depend on the nature of the contribution. For example
        \begin{enumerate}
            \item If the contribution is primarily a new algorithm, the paper should make it clear how to reproduce that algorithm.
            \item If the contribution is primarily a new model architecture, the paper should describe the architecture clearly and fully.
            \item If the contribution is a new model (e.g., a large language model), then there should either be a way to access this model for reproducing the results or a way to reproduce the model (e.g., with an open-source dataset or instructions for how to construct the dataset).
            \item We recognize that reproducibility may be tricky in some cases, in which case authors are welcome to describe the particular way they provide for reproducibility. In the case of closed-source models, it may be that access to the model is limited in some way (e.g., to registered users), but it should be possible for other researchers to have some path to reproducing or verifying the results.
        \end{enumerate}
    \end{itemize}

\item {\bf Open access to data and code}
    \item[] Question: Does the paper provide open access to the data and code, with sufficient instructions to faithfully reproduce the main experimental results, as described in supplemental material?
    \item[] Answer: \answerYes{}
    \item[] Justification: The HierFlow implementation and associated data-access instructions are provided through the accompanying code release, with algorithmic and experimental details documented in Sections~3--4 and Appendices~B--C.
    \item[] Guidelines:
    \begin{itemize}
        \item The answer \answerNA{} means that paper does not include experiments requiring code.
        \item Please see the NeurIPS code and data submission guidelines (\url{https://neurips.cc/public/guides/CodeSubmissionPolicy}) for more details.
        \item While we encourage the release of code and data, we understand that this might not be possible, so \answerNo{} is an acceptable answer. Papers cannot be rejected simply for not including code, unless this is central to the contribution (e.g., for a new open-source benchmark).
        \item The instructions should contain the exact command and environment needed to run to reproduce the results. See the NeurIPS code and data submission guidelines (\url{https://neurips.cc/public/guides/CodeSubmissionPolicy}) for more details.
        \item The authors should provide instructions on data access and preparation, including how to access the raw data, preprocessed data, intermediate data, and generated data, etc.
        \item The authors should provide scripts to reproduce all experimental results for the new proposed method and baselines. If only a subset of experiments are reproducible, they should state which ones are omitted from the script and why.
        \item At submission time, to preserve anonymity, the authors should release anonymized versions (if applicable).
        \item Providing as much information as possible in supplemental material (appended to the paper) is recommended, but including URLs to data and code is permitted.
    \end{itemize}

\item {\bf Experimental setting/details}
    \item[] Question: Does the paper specify all the training and test details (e.g., data splits, hyperparameters, how they were chosen, type of optimizer) necessary to understand the results?
    \item[] Answer: \answerYes{}
    \item[] Justification: Section~4.1 specifies the benchmarks, evaluation metrics, baselines, LLM oracle, and core hyperparameters, including candidate topology count, refinement budget, lower-level depth, branching factor, search budget, gating threshold, and three-run protocol. Appendix~C further gives dataset splits and baseline implementation details.
    \item[] Guidelines:
    \begin{itemize}
        \item The answer \answerNA{} means that the paper does not include experiments.
        \item The experimental setting should be presented in the core of the paper to a level of detail that is necessary to appreciate the results and make sense of them.
        \item The full details can be provided either with the code, in appendix, or as supplemental material.
    \end{itemize}

\item {\bf Experiment statistical significance}
    \item[] Question: Does the paper report error bars suitably and correctly defined or other appropriate information about the statistical significance of the experiments?
    \item[] Answer: \answerYes{}
    \item[] Justification: Table~2 reports mean performance with standard deviations across three independent runs for the main benchmark comparison, and Section~4.1 describes the experimental protocol. Appendix~C.6 additionally reports a statistically significant Spearman correlation for validating the gating proxy.
    \item[] Guidelines:
    \begin{itemize}
        \item The answer \answerNA{} means that the paper does not include experiments.
        \item The authors should answer \answerYes{} if the results are accompanied by error bars, confidence intervals, or statistical significance tests, at least for the experiments that support the main claims of the paper.
        \item The factors of variability that the error bars are capturing should be clearly stated (for example, train/test split, initialization, random drawing of some parameter, or overall run with given experimental conditions).
        \item The method for calculating the error bars should be explained (closed form formula, call to a library function, bootstrap, etc.)
        \item The assumptions made should be given (e.g., Normally distributed errors).
        \item It should be clear whether the error bar is the standard deviation or the standard error of the mean.
        \item It is OK to report 1-sigma error bars, but one should state it. The authors should preferably report a 2-sigma error bar than state that they have a 96\% CI, if the hypothesis of Normality of errors is not verified.
        \item For asymmetric distributions, the authors should be careful not to show in tables or figures symmetric error bars that would yield results that are out of range (e.g., negative error rates).
        \item If error bars are reported in tables or plots, the authors should explain in the text how they were calculated and reference the corresponding figures or tables in the text.
    \end{itemize}

\item {\bf Experiments compute resources}
    \item[] Question: For each experiment, does the paper provide sufficient information on the computer resources (type of compute workers, memory, time of execution) needed to reproduce the experiments?
    \item[] Answer: \answerNo{}
    \item[] Justification: The paper reports API cost and wall-clock time for efficiency analysis, but it does not provide complete compute-resource information such as compute worker type, CPU/GPU configuration, memory, storage, or total compute required for each experimental run.
    \item[] Guidelines:
    \begin{itemize}
        \item The answer \answerNA{} means that the paper does not include experiments.
        \item The paper should indicate the type of compute workers CPU or GPU, internal cluster, or cloud provider, including relevant memory and storage.
        \item The paper should provide the amount of compute required for each of the individual experimental runs as well as estimate the total compute. 
        \item The paper should disclose whether the full research project required more compute than the experiments reported in the paper (e.g., preliminary or failed experiments that didn't make it into the paper). 
    \end{itemize}
    
\item {\bf Code of ethics}
    \item[] Question: Does the research conducted in the paper conform, in every respect, with the NeurIPS Code of Ethics \url{https://neurips.cc/public/EthicsGuidelines}?
    \item[] Answer: \answerYes{}
    \item[] Justification: The research uses public benchmarks and LLM-based evaluation for an algorithmic study of workflow generation, and does not involve human subjects or sensitive private data. The work is intended to conform with the NeurIPS Code of Ethics.
    \item[] Guidelines:
    \begin{itemize}
        \item The answer \answerNA{} means that the authors have not reviewed the NeurIPS Code of Ethics.
        \item If the authors answer \answerNo, they should explain the special circumstances that require a deviation from the Code of Ethics.
        \item The authors should make sure to preserve anonymity (e.g., if there is a special consideration due to laws or regulations in their jurisdiction).
    \end{itemize}

\item {\bf Broader impacts}
    \item[] Question: Does the paper discuss both potential positive societal impacts and negative societal impacts of the work performed?
    \item[] Answer: \answerNo{}
    \item[] Justification: The current manuscript is primarily technical and does not include a dedicated discussion that covers both potential positive and negative societal impacts of automatic agentic workflow generation.
    \item[] Guidelines:
    \begin{itemize}
        \item The answer \answerNA{} means that there is no societal impact of the work performed.
        \item If the authors answer \answerNA{} or \answerNo, they should explain why their work has no societal impact or why the paper does not address societal impact.
        \item Examples of negative societal impacts include potential malicious or unintended uses (e.g., disinformation, generating fake profiles, surveillance), fairness considerations (e.g., deployment of technologies that could make decisions that unfairly impact specific groups), privacy considerations, and security considerations.
        \item The conference expects that many papers will be foundational research and not tied to particular applications, let alone deployments. However, if there is a direct path to any negative applications, the authors should point it out. For example, it is legitimate to point out that an improvement in the quality of generative models could be used to generate Deepfakes for disinformation. On the other hand, it is not needed to point out that a generic algorithm for optimizing neural networks could enable people to train models that generate Deepfakes faster.
        \item The authors should consider possible harms that could arise when the technology is being used as intended and functioning correctly, harms that could arise when the technology is being used as intended but gives incorrect results, and harms following from (intentional or unintentional) misuse of the technology.
        \item If there are negative societal impacts, the authors could also discuss possible mitigation strategies (e.g., gated release of models, providing defenses in addition to attacks, mechanisms for monitoring misuse, mechanisms to monitor how a system learns from feedback over time, improving the efficiency and accessibility of ML).
    \end{itemize}
    
\item {\bf Safeguards}
    \item[] Question: Does the paper describe safeguards that have been put in place for responsible release of data or models that have a high risk for misuse (e.g., pre-trained language models, image generators, or scraped datasets)?
    \item[] Answer: \answerNA{}
    \item[] Justification: The paper does not release a new pretrained language model, image generator, scraped dataset, or other high-risk artifact; it releases an implementation of the proposed method and evaluates on existing public benchmarks.
    \item[] Guidelines:
    \begin{itemize}
        \item The answer \answerNA{} means that the paper poses no such risks.
        \item Released models that have a high risk for misuse or dual-use should be released with necessary safeguards to allow for controlled use of the model, for example by requiring that users adhere to usage guidelines or restrictions to access the model or implementing safety filters. 
        \item Datasets that have been scraped from the Internet could pose safety risks. The authors should describe how they avoided releasing unsafe images.
        \item We recognize that providing effective safeguards is challenging, and many papers do not require this, but we encourage authors to take this into account and make a best faith effort.
    \end{itemize}

\item {\bf Licenses for existing assets}
    \item[] Question: Are the creators or original owners of assets (e.g., code, data, models), used in the paper, properly credited and are the license and terms of use explicitly mentioned and properly respected?
    \item[] Answer: \answerYes{}
    \item[] Justification: All datasets and models used are cited.
    \item[] Guidelines:
    \begin{itemize}
        \item The answer \answerNA{} means that the paper does not use existing assets.
        \item The authors should cite the original paper that produced the code package or dataset.
        \item The authors should state which version of the asset is used and, if possible, include a URL.
        \item The name of the license (e.g., CC-BY 4.0) should be included for each asset.
        \item For scraped data from a particular source (e.g., website), the copyright and terms of service of that source should be provided.
        \item If assets are released, the license, copyright information, and terms of use in the package should be provided. For popular datasets, \url{paperswithcode.com/datasets} has curated licenses for some datasets. Their licensing guide can help determine the license of a dataset.
        \item For existing datasets that are re-packaged, both the original license and the license of the derived asset (if it has changed) should be provided.
        \item If this information is not available online, the authors are encouraged to reach out to the asset's creators.
    \end{itemize}

\item {\bf New assets}
    \item[] Question: Are new assets introduced in the paper well documented and is the documentation provided alongside the assets?
    \item[] Answer: \answerYes{}
    \item[] Justification: The main new asset is the implementation of HierFlow. The method, prompt designs, hyperparameters, experimental protocol, and validation analyses are documented in Sections~3--4 and Appendices~B--C.
    \item[] Guidelines:
    \begin{itemize}
        \item The answer \answerNA{} means that the paper does not release new assets.
        \item Researchers should communicate the details of the dataset\slash code\slash model as part of their submissions via structured templates. This includes details about training, license, limitations, etc. 
        \item The paper should discuss whether and how consent was obtained from people whose asset is used.
        \item At submission time, remember to anonymize your assets (if applicable). You can either create an anonymized URL or include an anonymized zip file.
    \end{itemize}

\item {\bf Crowdsourcing and research with human subjects}
    \item[] Question: For crowdsourcing experiments and research with human subjects, does the paper include the full text of instructions given to participants and screenshots, if applicable, as well as details about compensation (if any)? 
    \item[] Answer: \answerNA{}
    \item[] Justification: The paper does not involve crowdsourcing experiments or research with human subjects; it evaluates automatic workflow generation on existing benchmarks.
    \item[] Guidelines:
    \begin{itemize}
        \item The answer \answerNA{} means that the paper does not involve crowdsourcing nor research with human subjects.
        \item Including this information in the supplemental material is fine, but if the main contribution of the paper involves human subjects, then as much detail as possible should be included in the main paper. 
        \item According to the NeurIPS Code of Ethics, workers involved in data collection, curation, or other labor should be paid at least the minimum wage in the country of the data collector. 
    \end{itemize}

\item {\bf Institutional review board (IRB) approvals or equivalent for research with human subjects}
    \item[] Question: Does the paper describe potential risks incurred by study participants, whether such risks were disclosed to the subjects, and whether Institutional Review Board (IRB) approvals (or an equivalent approval/review based on the requirements of your country or institution) were obtained?
    \item[] Answer: \answerNA{}
    \item[] Justification: The work does not involve crowdsourcing or human-subject studies, so IRB or equivalent approval is not applicable.
    \item[] Guidelines:
    \begin{itemize}
        \item The answer \answerNA{} means that the paper does not involve crowdsourcing nor research with human subjects.
        \item Depending on the country in which research is conducted, IRB approval (or equivalent) may be required for any human subjects research. If you obtained IRB approval, you should clearly state this in the paper. 
        \item We recognize that the procedures for this may vary significantly between institutions and locations, and we expect authors to adhere to the NeurIPS Code of Ethics and the guidelines for their institution. 
        \item For initial submissions, do not include any information that would break anonymity (if applicable), such as the institution conducting the review.
    \end{itemize}

\item {\bf Declaration of LLM usage}
    \item[] Question: Does the paper describe the usage of LLMs if it is an important, original, or non-standard component of the core methods in this research? Note that if the LLM is used only for writing, editing, or formatting purposes and does \emph{not} impact the core methodology, scientific rigor, or originality of the research, declaration is not required.
    \item[] Answer: \answerYes{}
    \item[] Justification: LLMs are a core methodological component of HierFlow, and the paper describes their use as initiator, refiner, optimizer, evaluator, and oracle components in Section~3 and Appendix~B. Section~4.1 further specifies that the main LLM oracle is instantiated with gpt-4o-mini for alignment with baselines.
    \item[] Guidelines:
    \begin{itemize}
        \item The answer \answerNA{} means that the core method development in this research does not involve LLMs as any important, original, or non-standard components.
        \item Please refer to our LLM policy in the NeurIPS handbook for what should or should not be described.
    \end{itemize}

\end{enumerate}

\end{document}